\documentclass[10pt,journal,compsoc]{IEEEtran}

\usepackage[nocompress]{cite}
\usepackage{booktabs} 

\usepackage{graphicx}
\usepackage{amsthm}
\usepackage{amsmath}
 \usepackage{amssymb}
\usepackage{adjustbox}

\usepackage{multirow} 

\usepackage{algorithmic}
\usepackage{algorithm}
\usepackage{textcase} 
\newtheorem{prop}{Proposition}
\usepackage{hyperref}

\usepackage{xcolor, eucal}

\newcommand{\eg}{\textit{e.g.}}
\newcommand{\ie}{\textit{i.e.}}
\DeclareMathOperator*{\argmin}{arg\,min}

\newcommand{\xvec}{{\mathbf{x}}}

\usepackage{hyperref}
\hypersetup{
    colorlinks=true,
    urlcolor=blue,
    linkcolor=blue,
    citecolor=blue,
    linkbordercolor=blue
}

\usepackage[margin=7pt,font=footnotesize,labelfont={sc,bf},labelsep=endash,tableposition=top]{caption}

\usepackage{mathtools}
\usepackage{bbm,nicefrac}

\begin{document}

\title{Explainable Deep Few-shot Anomaly Detection with Deviation Networks}
%
%
%

\author{Guansong Pang,
        Choubo   Ding,
        Chunhua  Shen, 
            Anton van den Hengel
\IEEEcompsocitemizethanks{\IEEEcompsocthanksitem 
G. Pang, C. Ding, C. Shen and A. van den Hengel are with University of Adelaide,
Australia.
%
}
\thanks{Manuscript received \today.}
}

\markboth{August 2021}%
{Pang \MakeLowercase{\textit{et al.}}: Explainable Deep Few-shot Anomaly Detection with Deviation Networks}

\IEEEtitleabstractindextext{%
\begin{abstract}

Existing anomaly detection paradigms overwhelmingly focus on training detection models using exclusively normal data or unlabeled data (mostly normal samples), assuming no access to any labeled anomaly data. One notorious issue with these approaches is that they are weak in discriminating anomalies from normal samples due to the lack of the knowledge about 
the anomalies. Here, we study the problem of few-shot anomaly detection, in which we aim at using a few labeled anomaly examples to train sample-efficient discriminative detection models. To address this problem, we introduce a novel weakly-supervised few-shot anomaly detection framework that can make use of the limited anomaly examples to train detection models without assuming the examples illustrating all possible classes of anomaly. 

Specifically, the proposed approach learns discriminative normality (regularity) by leveraging the labeled anomalies and a prior probability to enforce expressive representations of normality and unbounded deviated representations of abnormality. This is achieved by an end-to-end optimization of anomaly scores with a neural deviation learning, in which the anomaly scores of (anomaly-contaminated) normal samples are imposed to approximate scalar scores drawn from the prior while that of anomaly examples is 
enforced to have statistically significant deviations from these sampled scores in the upper tail. Furthermore, our model is optimized to learn fine-grained normality and abnormality by top-$K$ multiple-instance-learning-based feature subspace deviation learning, in which each data point is represented as a bag of multiple instances in different feature subspaces, allowing more generalized representations. Comprehensive experiments on nine real-world image anomaly detection benchmarks show that our model is substantially more sample-efficient and robust, and performs significantly better than state-of-the-art competing methods in both closed-set and open-set settings. Our model can also offer 
explanation capability as a result of its prior-driven anomaly score learning. Code and datasets are available at: \url{https://git.io/DevNet}.

\end{abstract}
}

\maketitle

\section{Introduction}

\begin{figure}[t!]
  \centering
    \includegraphics[width=0.485\textwidth]{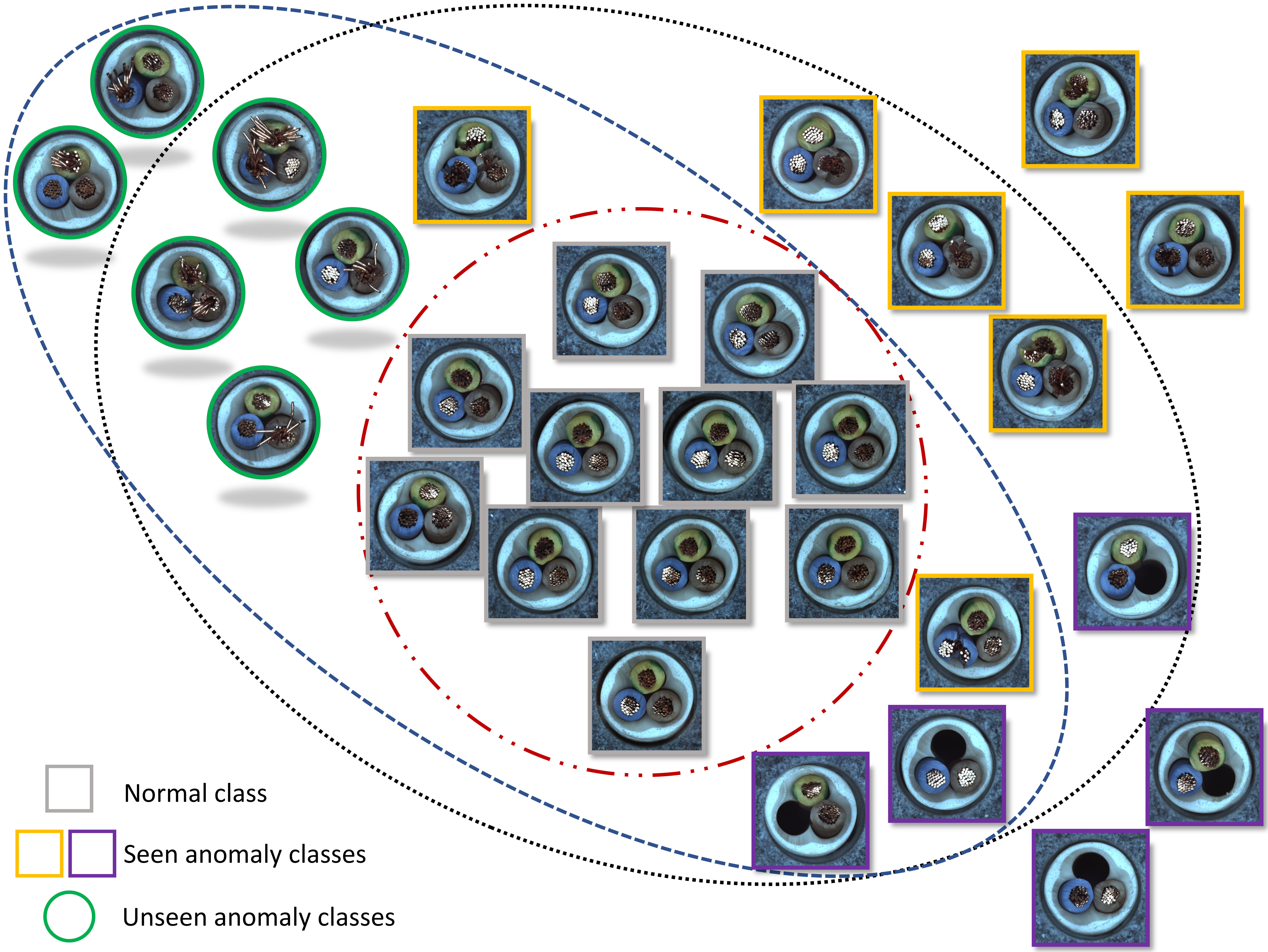}
  \caption{Cable defect detection examples of decision boundaries of three anomaly detection approaches: the unsupervised approach (black dashed line), fully-supervised approach (blue dashed line), and weakly-supervised approach (red dashed line). The weakly-supervised approach differs from the fully-supervised one in that i) it does not assume the given anomaly examples illustrating every possible class of anomaly; and ii) it does not assume large amount of labeled anomaly data. The purple and orange squares are cable defects -- `Missing Cable' and `Cut Inner Insulation', respectively. The green squares contain one type of unseen defect: `Bent Wire'.}
  \label{fig:motivationexample}
\end{figure}

Anomaly detection is the task of identifying data points that deviate significantly from the majority of data points,
which has a wide range of applications, 
such as  early diagnosing 
diseases using medical images or clinical data in healthcare, inspecting micro-cracks or defects on the surface of diverse objects in industrial inspection, detecting network intrusions in cybersecurity, and detecting fraudulent transactions in finance, among many others. There have been numerous anomaly detection methods introduced over the years, including both shallow and deep methods \cite{chandola2009anomaly,aggarwal2017outlieranalysis,pang2021deep}. Existing anomaly detection methods overwhelmingly focus on unsupervised/semi-supervised learning that trains detection models using exclusively normal data or unlabeled data that contains mainly normal samples, assuming no access to labeled anomaly data. The popularity of these methods is mainly to avoid the difficulty and/or prohibitive cost of collecting and manually labeling large-scale anomaly data in real-world applications. This type of approach learns the model of normal samples and detects any data points that deviate from the normal data as anomalies, as shown in Fig. \ref{fig:motivationexample}. One notorious issue with these methods is that they are weak in discriminating anomalies from normal samples due to the lack of the knowledge of the true anomalies, leading to high false positives/negatives \cite{gornitz2013toward,pang2018repen,siddiqui2018kdd,ruff2020deep,pang2021deep}. 

To tackle this issue, we study the problem of \textbf{few-shot anomaly detection} (\textbf{FSAD}), in which we aim at 
using 
a few labeled anomaly examples as some \textbf{partial} knowledge of the anomalies of interest to train sample-efficient and anomaly-informed detection models and reduce the false positives/negatives. The setting is viable since such anomaly knowledge is often available in many real-world anomaly detection applications. For instance, these labeled anomalies may originally come from a deployed detection system, \eg, a few successfully detected defect or network intrusion samples; or they may be from users/human experts, such as a small number of lesion images that are reported or confirmed by users/human experts. Due to the unbound nature and unknowingness of anomaly, FSAD is inherently a \textit{weakly-supervised learning} \cite{zhou2018brief} task, as the limited anomaly examples typically illustrate only an incomplete set of anomaly classes. Compared to fully-supervised anomaly detection that requires large-scale labeled anomaly data and works in a closed-set setting, FSAD aims to leverage the limited labeled anomaly data to learn well generalized detection models that can work effectively even in open-set scenarios where there are classes of anomaly unseen during training. As a result, FSAD models can detect both seen and unseen anomaly classes, while fully-supervised models fail to recognize unseen anomaly classes, as illustrated by the motivation example in Fig. \ref{fig:motivationexample}.

One major challenge in FSAD lies in learning generalized representations of abnormality using the limited anomaly examples. This is because: (i) the small labeled data may contain samples drawn from diverse anomaly classes of dissimilar class structures, which is different from general supervised learning tasks where samples of each class share the same class manifold, leading to great difficulty of learning a concrete representation of abnormality; and (ii) in addition to the seen anomalies, the learned abnormality representations need to generalize to unseen anomaly classes. Another major challenge is to learn robust representations of normality \textit{w.r.t.}\ the presence of potential anomaly contamination or data noises in the normal data. This is a ubiquitous challenge in most of the aforementioned applications where large-scale normal samples are readily accessible.

To address these two challenges, we introduce a novel weakly-supervised few-shot anomaly detection framework that utilizes the limited anomaly examples to train detection models without assuming the examples illustrating all possible classes of anomaly. The key idea is to train discriminative yet robust detection models by leveraging the labeled anomalies to learn prior-driven anomaly scores, enforcing robust representations of normality and unbounded deviated representations of abnormality. This is achieved by an end-to-end optimization of anomaly scores with a deviation learning based on a prior probability (\eg, Gaussian prior), in which the anomaly scores of (anomaly-contaminated) normal samples are imposed to approximate scalar scores drawn from the prior while that of anomaly examples are enforced to have statistically significant deviations from these sampled scores in the upper tail. The model is further optimized to obtain \textit{fine-grained} normality and abnormality by top-$K$ multiple-instance-learning-based feature subspace deviation learning, in which each data point is represented as a bag of multiple instances in different feature subspaces, resulting in more generalized representations.

\begin{figure}[t!]
  \centering
    \includegraphics[width=0.485\textwidth]{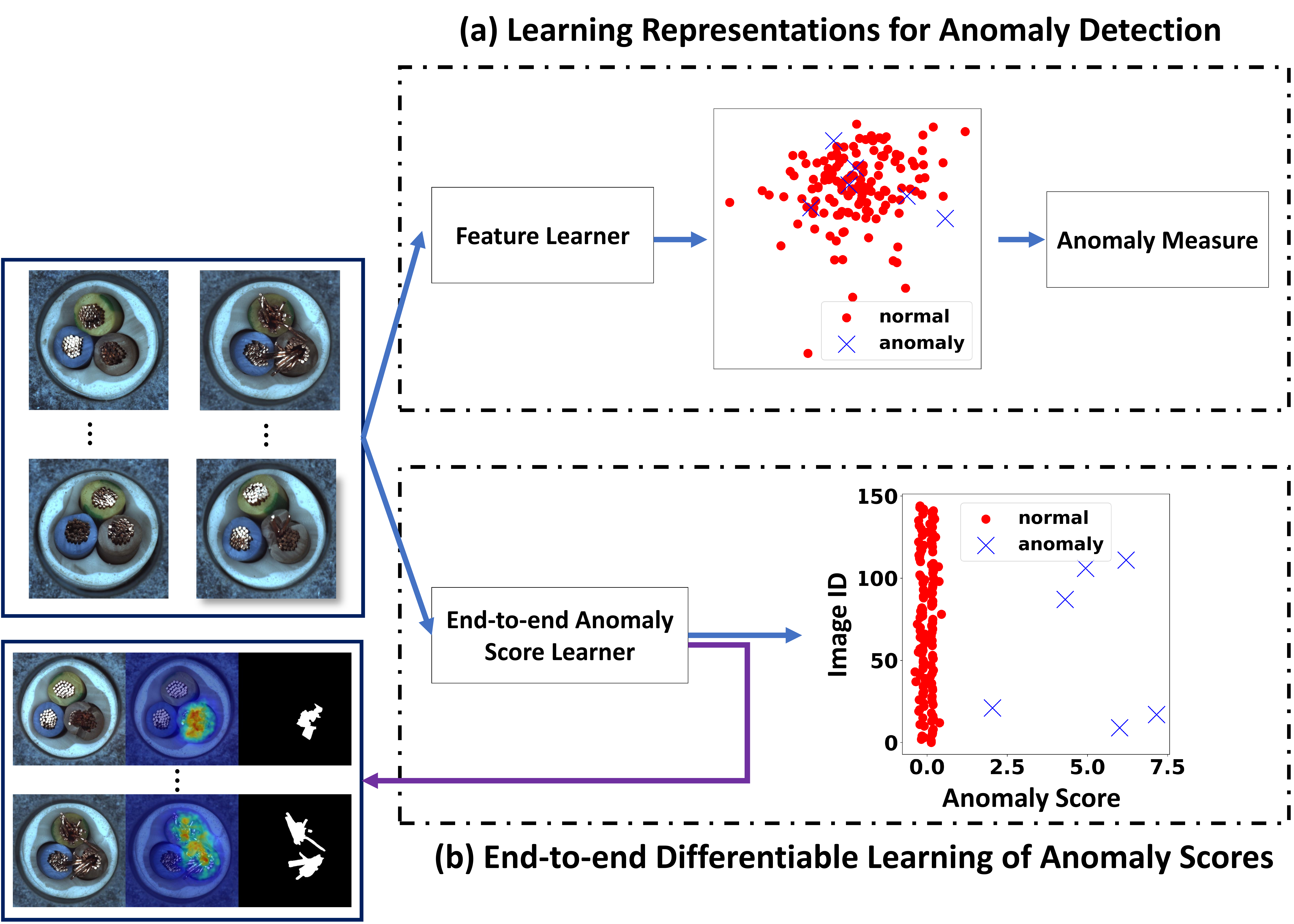}
  \caption{(a) Learning Features for Subsequent Anomaly Measures
vs. (b) Direct Learning of Anomaly Scores. It is easy for the latter approach to offer straightforward explanation about why specific data points are considered as anomalies by the deep models.}
  \label{fig:anomalyscorelearning}
\end{figure}

There have been a few explorations of FSAD in recent years, \eg, \cite{pang2018repen,ruff2020deep}, but they focus on optimizing the feature representations for some shallow anomaly scoring measures like one-class classification or distance measures, as illustrated in Fig. \ref{fig:anomalyscorelearning}(a), which is unfortunately an indirect optimization of anomaly scoring. This limits their exploitation of the anomaly examples, and their representation learning may also be restricted by the performance of their integrated anomaly measures. Further, they are difficult to obtain reliable explanation of the detected anomalies due to the weak coupling between the learned representations and the calculated anomaly scores. By contrast, as shown in Fig. \ref{fig:anomalyscorelearning}(b), our approach unifies representation learning and anomaly scoring into a single neural network, and directly optimizes the anomaly scores in an end-to-end fashion. As a result, our approach can fully exploit the anomaly examples to learn detection models with significantly improved generalizability and sample efficiency. The resulting model is also able to accurately localize and explain the cause of anomalies by attributing the anomaly scores to the inputs through gradient back-propagation.

To summarize, we make the following contributions.
\begin{itemize}
    \item We introduce a novel prior-driven anomaly detection framework to learn sample-efficient generalized detection models. In contrast to existing approaches that focus on feature learning, our framework fulfills a direct differentiable learning of anomaly scores. To our knowledge, this is the first framework for leveraging limited anomaly examples to achieve end-to-end anomaly score learning, resulting in significantly improved detection performance along with interpretable anomaly scores and accurate anomaly explanation.
    \item 
    A novel anomaly detection model, namely deviation networks (DevNet), 
    is instantiated from the
    proposed 
    framework. DevNet synthesizes Gaussian prior, Z-Score-based deviation loss, and multiple-instance learning to learn robust and well generalized models that can detect both seen and unseen classes of anomaly. 
    \item We present a theoretical analysis of the FSAD problem and show why DevNet can work effectively under open-set settings.
    \item We show, through extensive empirical results on nine real-world image datasets with real anomalies, that DevNet i) works effectively with different types of network backbones and achieves state-of-the-art performance in detecting different types of anomalies, \ie, texture and object anomalies in image data, significantly outperforming competing methods by a large margin in sample efficiency, robustness to anomaly contamination, and generalization to unseen anomalies; and ii) offers accurate anomaly localization for the explanation of the detected anomalies.
    \item We perform large-scale experiments to establish FSAD performance benchmarks on nine publicly available real-world datasets from diverse anomaly detection applications, offering baselines for the development and evaluation of FSAD methods.
    
\end{itemize}

\section{Related Work}

Here we review some work relevant to ours. 
\subsection{Anomaly Detection}

\subsubsection{Deep Learning-based Methods}
Most traditional anomaly detection approaches are ineffective in handling irrelevant features or non-linear feature relations in complex data \cite{chandola2009anomaly,aggarwal2017outlieranalysis,liu2012iforest,pang2018sparse}. There have been a large number of deep anomaly detection methods introduced in recent years to address these challenges \cite{pang2021deep}, including autoencoder-based approaches, GANs-based approaches, self-supervised approaches and  one-class classifiers, but they overwhelmingly focus on semi-supervised learning that trains detection models using exclusively normal data, or unsupervised learning from unlabeled data,
assuming no access to labeled anomaly data. 

Autoencoder-based approaches \cite{hawkins2002autoencoder,chen2017autoencoder,zhou2017autoencoder} use a bottleneck network architecture to learn a low-dimensional representation space, and then use the learned representations to define reconstruction errors as anomaly scores. GANs-based approaches \cite{schlegl2017gan,akcay2018ganomaly} also use the reconstruction error as an anomaly score, but they leverage two competing networks, a generator and a discriminator, to adversarially learn a latent space of the training data and use this latent space to compute the reconstruction errors. These deep methods can capture more complex feature interactions than traditional shallow methods, and they may be further enhanced by involving some memory modules to capture diverse normality patterns \cite{gong2019memorizing,park2020learning}. Self-supervised approaches \cite{golan2018deep,wang2019effective,bergman2020classification,georgescu2021anomaly} detect anomalies based on classification scores of a pre-text task of distinguishing the type of geometric/matrix transformation applied to the input data. One key issue with these three approaches is that their optimization objectives are often not primarily designed for anomaly detection, leading to suboptimal or unstable detection performance \cite{pang2018repen,ruff2018deepsvdd,ruff2020deep}. 

To address this issue, recent work \cite{pang2018repen,wang2020unsupervised,ruff2018deepsvdd,ruff2020deep,zong2018deep,chalapathy2018anomaly,wu2019deep,perera2019learning} focuses on coupling the representation learning objective with anomaly detection. For example, deep distance-based methods \cite{pang2018repen,wang2020unsupervised} integrate the representation learning with distance-based anomaly detectors, while deep one-class classifiers, such as deep support vector data description (SVDD) \cite{ruff2018deepsvdd,perera2019learning,ruff2020deep} and deep one-class SVM \cite{chalapathy2018anomaly,wu2019deep}, aim to learn representations for the one-class classification model. These approaches achieve large improvement over the previous methods. However, essentially their optimization objective still focuses on feature representations, so they optimize the anomaly scoring in an indirect manner. As shown by some recent work \cite{zenati2018gan,sabokrou2018adversarially,zaheer2020old,sabokrou2020deep,zheng2019one}, adversarial learning may be used in one-class classification to generate synthetic anomalies to achieve end-to-end differentiable learning of anomaly scores. Knowledge distillation \cite{bergmann2020uninformed,salehi2021multiresolution} has also shown effective in enhancing the feature learning for anomaly detection.

DevNet is different from all these approaches described above in that it performs a direct differentiable learning of the anomaly scores in an end-to-end fashion, without the reliance on the adversarial learning that is difficult to train and perform unstably \cite{zaheer2020old}. Further, they often have high false positives/negatives due to the lack of the knowledge of the true anomalies. DevNet addresses this issue by its weakly supervised anomaly-informed modeling.

\subsubsection{Few-shot Anomaly Detection}

Only a few studies 
perform 
anomaly detection with a few labeled anomalies. In \cite{mcglohon2009snare,tamersoy2014guilt}, a small set of labeled anomalies is incorporated into a belief propagation process to achieve more reliable anomaly scoring, but they are only applicable to graph data. In \cite{pang2018repen,liu2019margin,pang2019deep}, a few labeled anomalies are leveraged by deep metric learning to learn tailored feature representations for the distance/prediction error/regression-based anomaly detectors. Deep SAD \cite{ruff2020deep} utilizes a small set of anomaly examples and normal examples to boost the SVDD-based one-class classification when such information is available. These methods focus on feature representation learning only, while DevNet is focused on end-to-end anomaly score learning that can often leverage these limited labeled data more effectively. A deep reinforcement learning approach is presented in \cite{pang2021toward} to explore supervisory signals in large-scale unlabeled data while exploiting a small set of labeled anomalies. Additionally, multiple instance learning (MIL) has been explored in a number of studies \cite{sultani2018real,tian2021weakly} for weakly supervised anomaly detection where video-level class labels are given to detect frame-level anomalies, which is a different setting from ours. To our best knowledge, no results have been reported on leveraging MIL for few-shot anomaly detection.

A relevant but very different research line is to reduce the sample size of \textit{normal examples} in anomaly detection. Those research explores the adaptation of features learned in some relevant source domains to the target domain that has only a limited number of normal samples available \cite{lu2020few,reiss2021panda,sheynin2021hierarchical}.

\subsection{Anomaly Explanation}

Anomaly explanation is often as important as anomaly detection in many real-world applications. Previous work \cite{angiulli2009explanation,vinh2016explanation,gupta2018beyond} focuses on searching for anomalous features that can contrast the anomalies detected by some off-the-shelf anomaly detectors from a set of reference examples, so their detection and explanation are two decoupled stages. In deep anomaly detection, post-hoc interpretation techniques such as reconstruction error, attention mechanism, or gradient back-propagation \cite{chen2021unsupervised,pang2020ranking,venkataramanan2020attention,salehi2021multiresolution} are explored to generate saliency maps and provide faithful explanation about why specific examples are considered as anomalies. Similar to \cite{venkataramanan2020attention,pang2020ranking,salehi2021multiresolution}, the anomaly explanation module in DevNet is based on gradient back-propagation, but the end-to-end deviation learning enables DevNet to seamlessly offer more accurate and faithful anomalous feature localization for its detected anomalies (see Sec. \ref{subsec:explanation}).
 
\subsection{Learning 
with 
Limited Labeled Data}

Our studied problem is relevant to few-shot classification \cite{wang2020generalizing}, PU learning (\textit{i.e.}, learning from positive and unlabeled examples) \cite{bekker2020learning}, and imbalanced classification \cite{he2009imbalance,branco2016survey}. While they are relevant due to the objective of using a limited amount of labeled data to identify incoming examples of the class(es) of interest, FSAD is fundamentally different from these research lines. This is because i) they implicitly assume that the limited labeled examples and the test examples within the class of interest share the same manifold/class structure, whereas the few labeled anomalies themselves may be drawn from different anomaly classes, and further these labeled anomalies and the unseen test anomalies may be from different manifolds and exhibit entirely different feature expressions; and ii) anomaly detection is inherently an open-set task while these relevant tasks are typically formulated as a closed-set problem.

\section{The Proposed Framework}

\subsection{Problem Statement}

In addition to the access to large-scale (anomaly-contaminated) normal training data as assumed in most existing studies, the FSAD task also leverages a few labeled anomaly examples that may not illustrate all possible  classes  of  anomaly to learn anomaly-informed detection models. Specifically, given a set of $N+M$ training examples $\mathcal{X}=\{ \mathbf{x}_{1}, \mathbf{x}_{2}, \cdots, \mathbf{x}_{N}, \mathbf{x}_{N+1}, \mathbf{x}_{N+2}, \cdots, \mathbf{x}_{N+M} \}$
, in which $\mathcal{X}_n=\{ \mathbf{x}_{1}, \mathbf{x}_{2}, \cdots, \mathbf{x}_{N}\}$ is the normal data and $\mathcal{X}_a=\{\mathbf{x}_{N+1}, \mathbf{x}_{N+2}, \cdots, \mathbf{x}_{N+M} \}$ with $M\ll N$ is a very small set of labeled anomalies that provide some partial knowledge of true anomalies, our goal is to learn an anomaly scoring function $\phi: \mathcal{X} \rightarrow \mathbb{R}$ that assigns anomaly scores to data samples in a way that we have $\phi(\mathbf{x}_{i}) > \phi(\mathbf{x}_{j})$ if $\mathbf{x}_{i}$ is an anomaly and $\mathbf{x}_{j}$ is a normal example. Note that the normal training dataset $\mathcal{X}_n$ may be contaminated by some anomalies in practice, so $\phi$ is often required to be robust \textit{w.r.t.}\   such anomaly contamination.

For anomaly explanation, given a test example $\mathbf{x}$, let $\mathcal{F}$ be its full feature set, we aim to identify a feature subspace $\mathcal{S} \subset \mathcal{F}$ that the detection model $\phi$ relies on in determining whether $\mathbf{x}$ is an anomaly or not.

\subsection{Overview of Our Framework}

To solve this problem, we introduce a novel framework that synthesizes deep neural networks, a prior probability distribution of anomaly scores, and a new loss function to train an end-to-end differentiable anomaly scoring function, with an objective to assign statistically significantly larger anomaly scores to anomalies than normal examples. Due to the direct learning of anomaly scores, the resulting model is expected to yield more optimized anomaly scores and be more data-efficient than the current approaches whose objectives are to learn feature representations instead. 

\begin{figure*}[h!]
  \centering
    \includegraphics[width=0.95\textwidth]{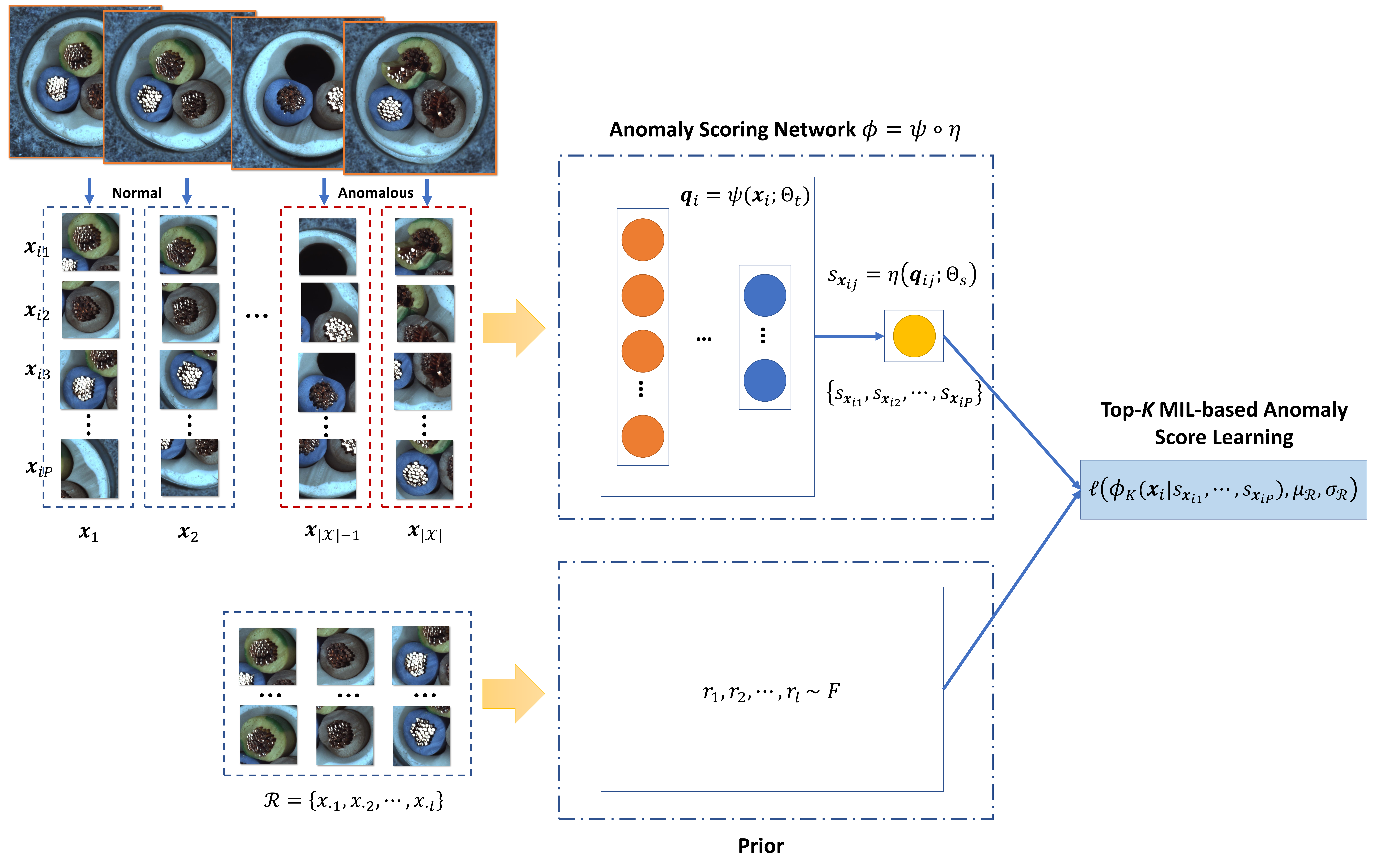}
  \caption{The Proposed Framework. $\phi$ is an anomaly score learner with the parameters $\Theta$. $\mu_{\mathcal{R}}$ is the mean of the anomaly scores of some normal examples, which is determined by a prior $F$. $\sigma_{\mathcal{R}}$ is a standard deviation associated with $\mu_{\mathcal{R}}$. $\phi_K$ is a top-$K$ multiple instance learning-based function for fine-grained anomaly scoring. The loss $\ell\big(\phi_K(\mathbf{x}), \mu_{\mathcal{R}}, \sigma_{\mathcal{R}}\big)$ is defined to guarantee that the anomaly scores of anomalies statistically significantly deviate from $\mu_{\mathcal{R}}$ in the upper tail while enforce normal examples have anomaly scores as close as possible to $\mu_{\mathcal{R}}$. }
  \label{fig:framework}
\end{figure*}

The procedure of our framework is shown in Fig. \ref{fig:framework}: 
\begin{enumerate}
    \item We first use an \textit{anomaly scoring network} (Sec. \ref{subsec:score_network}), \textit{i.e.}, a function $\phi$, to yield a scalar anomaly score for every image patch of a given image input $\mathbf{x}$.
    \item To guide the learning of anomaly scores, we then use a \textit{reference score generator} to generate a scalar score termed as \textit{reference score} (Sec. \ref{subsec:reference}), which is defined as the mean of the anomaly scores $\{r_1, r_2, \cdots, r_l\}$ for a set of $l$ randomly selected normal examples, denoted as $\mu_{\mathcal{R}}$. The reference score $\mu_{\mathcal{R}}$ is
    drawn from a prior probability $F$. This enables us to efficiently generate $\mu_{\mathcal{R}}$ and obtain interpretable anomaly scores. 
    \item Lastly, $\phi(\mathbf{x})$, $\mu_{\mathcal{R}}$ and its associated standard deviation $\sigma_{\mathcal{R}}$ are input to a  \textit{multiple-instance-learning (MIL)-based deviation loss} $\ell$ to guide the optimization (Sec. \ref{subsec:loss}), in which each image is represented as a bag of image patches and we aim to optimize the anomaly scores so that the scores of anomalous bags statistically significantly deviate from $\mu_{\mathcal{R}}$ in the upper tail while at the same time having the scores of normal examples as close as possible to $\mu_{\mathcal{R}}$. The use of the MIL-based deviation loss enables the model to learn more generalized representations than a holistic deviation loss, resulting in substantially reduced false detection errors (Sec. \ref{subsec:ablation}).
\end{enumerate}

The function $\phi$ is end-to-end optimized by minimizing the proposed deviation loss $\ell$ during training. At the inference stage, given a test image, $\phi$ directly yields an anomaly score, which can be further back-propagated to raw image pixels to infer which image pixels are the main contributors to the abnormality for the anomaly explanation.
 
\subsection{Key Intuition}

The deviation loss-based optimization in our framework forces the normal examples to cluster around $F$ in terms of their anomaly scores but pushes anomalies statistically far away $F$, which well optimizes the anomaly scores and also empowers the intermediate representation learning to discriminate normal examples from the rare anomalies with different anomalous behaviors. In other words, our deep anomaly detector leverages a few labeled anomalies and the prior of anomaly scores to learn a high-level abstraction of normality. Further, the MIL-based deviation learning enforces the normality learning at fine-grained image patch level, which effectively eliminates the interference of normal patches in the anomalous images and facilitates the learning of more generalized normal features and deviated abnormal features (see Sec. \ref{subsec:fsad_analysis}). The resulting model would assign a large anomaly score to a test example as long as its features significantly deviate from the learned normality abstraction. This offers an effective detection of dissimilar anomalies, \eg, previously unseen anomalies; and in turn the differentiable anomaly score learning also requires substantially less labeled anomalies to train the detector.

\section{Deviation Networks}

The proposed framework is instantiated into a model called Deviation Networks (DevNet), which defines a Gaussian prior and a MIL-driven Z-Score-based deviation loss to enable the differentiable learning of anomaly scores, enabling the sample-efficient learning of generalized representations.

\subsection{End-to-end Anomaly Scoring Network}\label{subsec:score_network}
Different from most existing studies that learn a feature representation mapping, DevNet directly learns a anomaly score mapping $\phi$. Let $\mathcal{Q} \in \mathbb{R}^{L}$ be an intermediate representation space, DevNet devises an anomaly scoring network $\phi(\cdot; \Theta):\mathcal{X} \rightarrow \mathbb{R}$ composed of a feature representation learner $\psi(\cdot; \Theta_{t}): \mathcal{X} \rightarrow \mathcal{Q}$ and an anomaly scoring function $\eta(\cdot; \Theta_{s}): \mathcal{Q} \rightarrow \mathbb{R}$, in which $\Theta=\{\Theta_{t}, \Theta_{s}\}$. Particularly, let $\mathcal{B}(\mathbf{x}_{i})$ be a bag of image patches derived from the image $\mathbf{x}$, then $\psi(\cdot; \Theta_{t})$ is defined to be a \textit{fine-grained feature learner} with $H \in \mathbb{N}$ hidden layers and their weight matrices $\Theta_{t}=\{\mathbf{W}^{1}, \mathbf{W}^{2}, \cdots, \mathbf{W}^{H}\}$, which can be represented as
\begin{equation}
    \mathbf{q}_{ij} = \psi(\mathbf{x}_{ij}; \Theta_{t}),
\end{equation}
where $\mathbf{x}_{ij} \in \mathcal{B}(\mathbf{x}_{i})$ and $\mathbf{q}_{ij} \in \mathcal{Q}$. Different hidden network architecture can be used here based on the type of data inputs, such as multilayer perceptron networks for multidimensional data, convolutional networks for image data, or recurrent networks for sequence data. This work focuses on image data. ResNet-18 \cite{he2016deep} is used to instantiate $\psi$ by default; DevNet can also perform well with other convolutional network architectures, as shown in Sec. \ref{subsec:ablation}.

To directly learn the anomaly scores, an \textit{anomaly score learner} $\eta(\cdot, \Theta_{s}): \mathcal{Q} \rightarrow \mathbb{R}$ is further defined, which uses a single linear neural unit in the output layer to compute the anomaly scores based on the intermediate representations:
\begin{equation}
    \eta(\mathbf{q}_{ij};\Theta_{s}) = \sum_{k=1}^{L}w^{o}_{k} q_{ijk} + w^{o}_{L+1},
\end{equation}
where $q_{ijk}$ is an entry of the feature vector $\mathbf{q}_{ij}$ and $\Theta_{s} = \{\mathbf{w}^{o}\} $ ($w^{o}_{L+1}$ is the bias term).

Thus, $\phi(\cdot; \Theta)$ can be formally represented as 
\begin{equation}
    \phi(\mathbf{x}_{ij};\Theta) = \eta(\psi(\mathbf{x}_{ij};\Theta_{t});\Theta_s),
\end{equation}
which directly maps data inputs to scalar anomaly scores and can be trained in an end-to-end fashion.

\subsection{Gaussian Prior-based Reference Scores}\label{subsec:reference}

Having obtained the anomaly scores using $\phi(\mathbf{x};\Theta)$, a key problem is to provide some type of knowledge to guide the optimization of the anomaly scores. To address this problem, DevNet devises a module to generate a \textit{reference score} $\mu_{\mathcal{R}} \in \mathbb{R}$, which is defined as the mean of the anomaly scores of a set of some randomly selected normal examples $\mathcal{R}$. This score is directly fed into the network output to guide the optimization of $\phi$. There are two main ways to generate $\mu_{\mathcal{R}}$: data-driven and prior-driven approaches. Data-driven methods involve a model to learn $\mu_{\mathcal{R}}$ based on $\mathcal{X}$, while prior-driven methods generate $\mu_{\mathcal{R}}$ from a chosen prior probability $F$. The prior-based approach is chosen here because (i) the chosen prior allows us to achieve good interpretability of the predicted anomaly scores and (ii) it can generate $\mu_{\mathcal{R}}$ constantly, which is substantially more efficient than the data-driven approach.

The specification of the prior is the main challenge of the prior-based approach. Fortunately, extensive results in \cite{kriegel2011interpreting} show that Gaussian distribution fits the anomaly scores very well in a range of datasets. This is partly due to that the most general distribution for fitting values derived from Gaussian or non-Gaussian variables is the Gaussian distribution according to the central limit theorem. Motivated by this, we define a Gaussian prior-based reference score:
\begin{align}
    r_1, r_2, \cdots, r_l & \sim {\cal N}  (\mu, \sigma^{2}), \\
    \mu_{\mathcal{R}} & = \frac{1}{l}\sum_{i=1}^{l}r_i,
\end{align}
where each $r_i$ is drawn from ${\cal N} (\mu, \sigma^{2})$ and represents an anomaly score of a random normal data example. 
$\mu=0$ and $\sigma=1$ are used in our experiments, which help DevNet to achieve stable detection performance on different datasets. DevNet is not sensitive to $l$ when $l$ is sufficiently large due to the central limit theorem. $l=5000$ is used here.

\subsection{MIL-driven Deviation Loss}\label{subsec:loss}

A deviation loss is further defined to utilize the Gaussian prior-based scores to optimize the anomaly scoring network. Particularly, the deviation is specified as a Z-Score as follows:
\begin{equation}\label{eqn:mil}
    \phi_{K}(\mathbf{x}_i;\Theta) = \max_{\mathcal{M}(\mathbf{x}_i) \subset \mathcal{B}(\mathbf{x}_i), |\mathcal{M}(\mathbf{x}_i)|=K} \frac{1}{K} \sum_{\mathbf{x}_{ij} \in \mathcal{M}(\mathbf{x}_i)} \phi(\mathbf{x}_{ij};\Theta),
\end{equation}
\begin{equation}\label{eqn:deviation}
    \mathit{dev}(\mathbf{x}_{i};\Theta) = \frac{\phi_{K}(\mathbf{x}_{i};\Theta) - \mu_{\mathcal{R}}}{\sigma_{\mathcal{R}}},
\end{equation}
where $\mathcal{M}(\mathbf{x}_i)$ is a set of image patches that have the largest anomaly scores among all patches of $\mathbf{x}_i$, and $\sigma_{\mathcal{R}}$ is the standard deviation of the prior-based anomaly score set, $\{r_1, r_2, \cdots, r_l\}$. The top-$K$ anomaly score selection and aggregation in Eqn.~(\ref{eqn:mil}) is motivated by the following two observations: i) there are often some normal/background regions for each anomalous image and ii) there are some `hard' regions within normal images that are difficult to be distinguished from the anomalous image patches. Thus, learning with only the top-$K$ (\ie, $K=|\mathcal{M}(\mathbf{x}_i)|$) anomalous patches per image enables i) a more reliable learning of anomaly scores for anomaly images and ii) the selection of `hard' negative samples for the scoring optimization.

The deviation is lastly plugged into the contrastive loss \cite{hadsell2006contrastloss} to specify our deviation loss to train $\phi_K$.
\begin{multline}\label{eqn:loss}
    \ell\big(\mathbf{x}_{i}, \mu_{\mathcal{R}}, \sigma_{\mathcal{R}};\Theta \big) = (1-y_{i})|\mathit{dev}(\mathbf{x}_{i};\Theta)|\\ + y_{i} \max\big(0, a - \mathit{dev}(\mathbf{x}_{i};\Theta)\big),
\end{multline}
where $y_{i}=1$ if $\mathbf{x}_{i} \in \mathcal{X}_a$ and $y_{i}=0$ if $\mathbf{x}_{i} \in \mathcal{X}_u$, and $a$ is equivalent to a Z-Score confidence interval parameter. This loss enables DevNet to push the anomaly scores of normal examples as close as possible to $\mu_{\mathcal{R}}$ while enforcing a deviation of at least $a$ between $\mu_{\mathcal{R}}$ and the anomaly scores of the anomalies. Note that if $\mathbf{x}_{i}$ is an anomaly, and it has a negative $\mathit{dev}(\mathbf{x}_{i})$, the loss is particularly large, which encourages large \textit{positive deviations} for all anomalies. Therefore, the deviation loss is equivalent to enforcing a statistically significant deviation of the anomaly score of all anomalies from that of normal examples in the upper tail. We use $a=5$ to achieve a very high significance level (\textit{i.e.}, 
$5.73303\cdot 10^{-7}$%
) for all labeled anomalies. 

Similar to the contrastive loss, the deviation loss is monotonically increasing in its first term, $|\mathit{dev}(\mathbf{x}_{i})|$, and is monotonically deceasing in the second term, $\max\big(0, a - \mathit{dev}(\mathbf{x}_{i})\big)$, so it is convex \textit{w.r.t.}\   both cases. However, they are also very different, because the contrastive loss uses pairs of intra-class/inter-class data examples as training samples to learn a similarity metric, whereas our deviation loss is built upon the deviation function and dedicated to the differentiable learning of fine-grained anomaly scores.

\subsection{Back-propagation-based Anomaly Explanation}\label{subsec:explanation_method} 

Since DevNet is trained in an end-to-end manner to learn the anomaly scores, the contribution of each image feature to its output anomaly score can be directly inferred by the derivative of the output score \textit{w.r.t.} the input image features \cite{selvaraju2017grad}. The explainability of an anomaly detector is quantified by the success of locating anomaly features that should in turn contributes most to the anomaly score. 
Motivated by the success of Grad-CAM-based methods in many tasks \cite{du2019techniques}, a gradient-based back propagation method is used in DevNet for its anomaly localization. Particularly, for a given image $\mathbf{x}_i$, the contribution of each image feature $x_{ij}$ to the anomaly score is defined by its gradient:
\begin{equation}\label{eqn:localization}
M_{ij}=g_\sigma (\left | \frac{\partial \phi_{K}(\mathbf{x}_i;\Theta^*)}{\partial x_{ij}} \right |),
\end{equation}
where $\mathbf{M}$ is a saliency map with each entry $M_{ij}$ corresponding to the contribution of a specific image feature to the overall anomaly score yielded by DevNet, $g_{\sigma}$ is a Gaussian filter with a standard deviation $\sigma$, which is used to blur the saliency map to eliminate noise and obtain more smooth anomaly localization, and $\Theta^*$ are the parameters learned after optimizing DevNet. In our experiments, a 16x16 sliding Gaussian kernel with $\sigma=4$ is applied to the saliency map to obtain the anomaly localization map.

\subsection{The DevNet Algorithm}\label{subsec:algo}

Algorithm \ref{alg:devnet} presents the procedure of training DevNet. After random weight initialization, DevNet performs stochastic gradient descent-based optimization to learn the weights in $\Theta$ in Steps 2-10. Particularly, Step 4 first samples a mini-batch $\mathcal{B}$ of size $b$ using stratified random sampling, followed by sampling the anomaly scores of $l$ normal examples from the prior $\mathcal{N}(\mu, \sigma^2)$ in Step 5. After obtaining $\mu_{\mathcal{R}}$ and $\sigma_{\mathcal{R}}$ in Step 6, Step 7 performs the forward propagation of the anomaly scoring network and computes the loss. Step 8 then performs gradient descent steps \textit{w.r.t.} the parameters in $\Theta$. We finally obtain the optimized scoring network $\phi$.

\renewcommand{\algorithmicrequire}{\textbf{Input:}}
\renewcommand{\algorithmicensure}{\textbf{Output:}}
\begin{algorithm}[t]
\small 
\caption{\textit{Training DevNet}}
\begin{algorithmic}[1]
\label{alg:devnet}
\REQUIRE $\mathcal{X} \in \mathbb{R}^{D}$ - training data examples
\ENSURE $\phi: \mathcal{X} \rightarrow \mathbb{R}$ - an anomaly scoring network
\STATE Randomly initialize $\Theta$ in $\phi$
\FOR{ $i = 1$ to $\mathit{n\_epochs}$}
    \FOR{ $j = 1$ to $\mathit{n\_batches}$}
        \STATE $\mathcal{B} \leftarrow$ Randomly sample $\mathit{b}$ data examples with a half of examples from $\mathcal{X}_a$ and another half from $\mathcal{X}_n$
        \STATE Randomly sample $l$ anomaly scores from $\mathcal{N}(\mu, \sigma^2)$
        \STATE Compute $\mu_{\mathcal{R}}$ and $\sigma_{\mathcal{R}}$ of the $l$ sampled anomaly scores
        \STATE $\mathit{loss} \leftarrow \frac{1}{b}\sum_{\mathbf{x} \in \mathcal{B}}\ell\big(\mathbf{x}, \mu_{\mathcal{R}}, \sigma_{\mathcal{R}};\Theta \big)$    
        \STATE Perform a gradient descent step \textit{w.r.t.} the parameters in $\Theta$
    \ENDFOR
\ENDFOR
\RETURN $\phi$ with the learned parameters $\Theta^*$
\end{algorithmic}
\end{algorithm}

At the testing stage, DevNet uses the optimized $\phi$ and Eqn. (\ref{eqn:deviation}) to produce an anomaly score for every test example and returns an anomaly ranking of the data examples based on the anomaly scores, in which the top-ranked examples are anomalies. Most existing methods also output the anomaly scores, rather than binary class labels \cite{chandola2009anomaly,pang2021deep}. Different from these studies whose anomaly scores are often hard to be interpreted \cite{kriegel2011interpreting}, our anomaly scores are highly interpretable due to the Gaussian prior used in the reference score generation (Sec. \ref{subsec:interpretability_score}). Further, the output anomaly scores can be directly used to locate anomaly features via Eqn. (\ref{eqn:localization}) to provide anomaly explanation.

\section{Theoretical Analysis of DevNet}

\subsection{Anomaly Score Learning under the FSAD Setting}\label{subsec:fsad_analysis}

We show here that top-$K$ MIL-based anomaly score learning is more plausible than the holistic image-based deviation learning under the FSAD Setting. Specifically, FSAD can be formulated as a \textit{supervised open-set one-class classification} problem. It is a supervised one-class classification because both labeled anomaly and normal examples are used in learning the decision function for recognizing the normal class; it is an open-set one-class classification in that the labeled anomaly examples do not illustrate every class of anomaly, \ie, there is some open space $\mathcal{O}$ that lies outside of the labeled normal training examples in $\mathcal{X}_n$ and the labeled anomaly examples in $\mathcal{X}_a$. Particularly, let $f(\xvec) \in \mathcal{H}$ be a function from a fixed class of one-class classification functions where $f(\xvec)=1$ indicates the identification of $\xvec$ as normal and $f(\xvec)=0$ when $\xvec$ is considered as anomaly, following \cite{scheirer2012toward}, the FSAD problem is to find $f^*$ such that
\begin{equation}\label{eqn:bothrisk}
    f^* = \argmin_{f \in \mathcal{H}}\{R_{\mathcal{O}}(f) + \lambda R_{\mathcal{E}}\big(f(\mathcal{X}_a \cup \mathcal{X}_u)\big) \},
\end{equation}
where $R_{\mathcal{E}}=\mathbb{E}\big[\ell(f(\xvec),y)\big]$ is an empirical risk defined over the labeled training examples, $\lambda$ is a regularization term, and $R_{\mathcal{O}}$ is an open space risk that is defined as the fraction of open space classified as normal compared to the overall space classified as normal:
\begin{equation}\label{eqn:openrisk}
    R_{\mathcal{O}}(f)=\frac{\int_{\mathcal{O}}f(\xvec)d\xvec}{\int_{\mathcal{R}}f(\xvec)d\xvec},
\end{equation}
where $\mathcal{R}$ is a large space that contains all the normal training examples in $\mathcal{X}_n$ and an open space $\mathcal{O}$ that is labeled as normal. Labeling more open space as normal indicates a greater open space risk. Thus, the key to FSAD is to exploit the limited anomaly examples to minimize the open space risk $R_{\mathcal{O}}$, in addition to the minimization of the empirical risk $R_{\mathcal{E}}$.

DevNet uses the deviation loss in Eqn. (\ref{eqn:loss}) to minimize the empirical risk over the labeled data. To reduce the open space risk, DevNet increases the denominator term $\int_{\mathcal{R}}f(\xvec)d\xvec$ in Eqn. (\ref{eqn:openrisk}) via the top-$K$ MIL anomaly score learning in Eqn. (\ref{eqn:mil}). Particularly, when minimizing a holistic image-based deviation loss that treats the entire region of each anomaly example as anomalous, DevNet can be inclined to the anomaly examples as the term $\max\big(0, a - \mathit{dev}(\mathbf{x}_{i};\Theta)\big)$ can incur greater loss than $|\mathit{dev}(\mathbf{x}_{i};\Theta)|$. One issue here is that the labeled anomaly example set $\mathcal{X}_a$ typically contains a large proportion of normal image patches that also appear in the normal example set $\mathcal{X}_n$. Consequently, the holistic image-based deviation loss wrongly treats the normal image patches in $\mathcal{X}_a$ as abnormal, which can suppress the learned normal space (\ie, smaller $\int_{\mathcal{R}}f(\xvec)d\xvec$ due to the suppression of $\int_{\mathcal{R}-\mathcal{O}}f(\xvec)d\xvec$).
By contrast, the top-$K$ MIL anomaly score function $\phi_K(\xvec)$ instead learns the anomaly score using only the most likely anomalous image patches in each anomaly example. This helps avoid the overfitting to $\mathcal{X}_a$ and reduce the suppression of the normal space, resulting in larger $\int_{\mathcal{R}-\mathcal{O}}f(\xvec)d\xvec$ (and thus larger $\int_{\mathcal{R}}f(\xvec)d\xvec$) compared to the holistic image-based approach (See Sec. \ref{subsec:ablation} for empirical justification).

\subsection{Interpretability of Anomaly Scores} \label{subsec:interpretability_score}
Given the anomaly scores yielded by most anomaly detectors, it is not clear what is the probability of each test example being an anomaly, and it is also difficult to determine a specific threshold to select the top-ranked examples \cite{kriegel2011interpreting}. Therefore, if users need more than an anomaly ranking in practice, some types of separate anomaly score unification methods \cite{kriegel2011interpreting} are required for those methods to transform their scores into more interpretable ones. However, the anomaly scoring and the score unification are two independent modules in such cases, which may lead to untrustworthy explanation of the scores. By contrast, DevNet directly yields interpretable anomaly scores.

\begin{prop}
Let $\mathbf{x} \in \mathcal{X}$ and $z_{p}$ be the quantile function of $\mathcal{N}(\mu, \sigma^{2})$, then $\phi_K(\mathbf{x})$ lies outside the interval $\mu \pm z_{p}\sigma $ with a probability $2(1-p)$.
\end{prop}

This proposition of DevNet is due to the Gaussian prior and Z-Score-based deviation loss. The probability $2(1-p)$ offers a straightforward explanation to the anomalousness of any given score $\phi_K(\mathbf{x})$. Particularly, we have the probability $(1-p)$ when only focusing on the upper tail $\mu + z_{p}\sigma $, \textit{e.g.}, by applying $p=0.95$, we have $z_{0.95}=1.96$, which states that having anomaly scores over 1.96 (as $\mu=0$ and $ \sigma=1 $ are used in DevNet) indicates that the example only has a probability of 0.05 generated from the same mechanism as the normal class. Users can also easily choose a threshold to determine anomalies with a desired confidence level, \textit{e.g.}, given the anomaly score distribution shown in Fig. \ref{fig:anomalyscorelearning}(b), it is easy to use $z_{0.95}$ to identify the anomalies with a 95\% confidence level.

\section{Experiments}

Diverse practical experiment settings are used to facilitate a comprehensive evaluation of DevNet to answer the following six key questions.

\begin{itemize}
    \item \textbf{Q1. How effective does DevNet leverage a few random labeled anomalies?} We evaluate the performance on many real-world application settings where a very small number of labeled anomalies randomly sampled from an unknown anomaly class distribution are available during training, in addition to the access to large normal samples.
    \item \textbf{Q2. How does DevNet generalize to open-set anomaly detection?} Through this question, we aim to explicitly evaluate the performance of DevNet under the open-set setting, in which all the anomaly classes in the test data are different from the classes of labeled anomalies available during training. 
    \item \textbf{Q3. How is the explainability of the detection model?} This question examines the effectiveness of detection models in providing accurate causes of anomalies, \ie, locating the abnormal region of an anomaly image.
    \item \textbf{Q4. What is the contribution of each module of DevNet to its overall performance?} We also evaluate the performance of each module of DevNet via comprehensive ablation study.
    \item \textbf{Q5. How is the sample efficiency of DevNet?} This question examines the performance of DevNet \textit{w.r.t.}\  different amount of labeled anomaly data.
    \item \textbf{Q6. How robust is DevNet \textit{w.r.t.}\  different levels of anomaly contamination?} Training normal data is often contaminated by some anomaly data in many applications, so we evaluate the robustness of the detectors \textit{w.r.t.}\  different anomaly contamination rates.
\end{itemize}

\subsection{Datasets}
Nine publicly available image datasets with real anomalies from diverse application domains are used. These include five benchmarks for identifying visual defects or micro-cracks on different object surfaces: MVTec AD\footnote{\url{https://tinyurl.com/mvtecad}}, AITEX\footnote{\url{https://tinyurl.com/aitex-defect}}, SDD\footnote{\url{https://tinyurl.com/KolektorSDD}}, ELPV\footnote{\url{https://tinyurl.com/elpv-crack}}, and Optical\footnote{\url{https://tinyurl.com/optical-defect}}; a benchmark for detecting novel observations from large-scale observations on the surface of planets (such as Mars) in rover-based planetary exploration: Mastcam\footnote{\url{https://tinyurl.com/mastcam}}; and three medical imaging benchmarks for detecting lesions on colonoscopy/MRI/CT images: Hyper-Kvasir\footnote{\url{https://tinyurl.com/hyper-kvasir}}, BrainMRI\footnote{\url{https://tinyurl.com/brainMRI-tumor}}, and HeadCT\footnote{\url{https://tinyurl.com/headCT-tumor}}. 

\begin{itemize}
    \item \textbf{MVTec AD} \cite{bergmann2019mvtec} is a real-world industrial defect inspection dataset that contains 5,354 high-resolution images covering five types of texture defects and ten types of object defects. Each type of defect contains one to eight fine-grained types of defects. In total, there are 73 types of anomalies in the form of texture/object-level defects.
    \item \textbf{AITEX} \cite{silvestre2019public} consists of 245 4,096x256 images of 7 different fabrics. There are 140 defect-free images, 20 for each type of fabric, and 105 defect images of 12 defect classes. Each 4096x256 image is cropped to create 16 256x256 patch-based images, with each patch image relabeled, \ie, it is a defect sample if it contains any part of the defects, and it is a non-defect sample otherwise. This results in 2,256 non-defect images and 183 defect images.
    \item \textbf{SDD} \cite{Tabernik2019JIM} contains 399 images with a resolution of 500x1240, including 52 defect images and 347 normal images. Each image is vertically and evenly splitted into three segment images, with each segment image relabeled. This results in 880 normal images and 54 defect images.
    
    \item \textbf{ELPV} \cite{deitsch2019elpv} contains 2,624 300x300 images of functional and defective solar cells captured from 44 different solar modules. The defects are manually annotated by domain experts and categorized into two classes: mono- and poly-crystalline according to the type of solar modules. The mono-crystalline class has 313 defective images while the poly-crystalline class contains 403 defective images.

    \item \textbf{Optical} \cite{wieler2007weakly} is a dataset used to simulate industrial optical inspection. It contains 16,100 512x512 images, of which 
    14,000 images have no defects and 2,100 defect images. 

    \item \textbf{Mastcam} \cite{kerner2020comparison} is a Mars geological image datasets collected by the Mastcam imaging system, including 9,728 64x64x6 Mars geological images. Each image includes three color channels and three grayscale channels. We focus on three-channel images in our experiments, so only the three color channels are used. It contains 451 expert-identified geologically-interesting/novel images, which are categories into 11 classes: `meteorite', `float rock', `bedrock', `vein', `broken rock', `dump pile', `drill hole', `dust removal tool (DRT) spot', `scuff', `edge cases', and `other'. 

    \item \textbf{BrainMRI} \cite{salehi2021multiresolution} is a medical dataset consisting of 253 brain MRI images, for which images with tumors are treated as anomalies (155 images) while the healthy ones are considered as normal (98 images). 

    \item \textbf{HeadCT} \cite{salehi2021multiresolution}
    includes 100 normal head computed tomography (CT) images and 100 images with hemorrhage. The hemorrhage images are considered as anomalies.

    \item \textbf{Hyper-Kvasir} \cite{borgli2020hyperkvasir} is the largest gastrointestinal dataset, which is collected during real gastro and colonoscopy examinations at a Norway hospital. It contains 10,662 images of 23 different classes manually labeled by experienced gastrointestinal endoscopists. We use 3,452 upper gastrointestinal tract images,
    with the anatomical landmark category as normal class and the pathological category as abnormal class. The pathological category includes four subclasses, \ie, barretts (41 images), barretts-short-segment (53 images), esophagitis-a (403 images), and esophagitis-b-d (260 images).
    
\end{itemize}

Many anomaly detection methods are evaluated on image classification benchmarks, such as MNIST, CIFAR-10, CIFAR-100, by using synthetic anomaly detection protocols, \eg, one-vs-all protocol \cite{ruff2018deepsvdd}. However, these datasets may fail to replicate the complexities of real-world anomaly detection. For example, as shown by a number of recent studies \cite{bergmann2020uninformed,yi2020patch,salehi2021multiresolution}, the methods that achieve excellent performance (AUC-ROC of 90+\%) on those datasets can work ineffectively on datasets with real anomalies (\eg, AUC-ROC of 60\%-70\% on MVTec AD). Additionally, there are also a number of video anomaly detection benchmarks that may be converted for image anomaly detection, \eg, by treating each video frame as an image sample. However, this can lead to a significant loss of temporal information that is crucial for detecting abnormal events in videos. Motivated by these observations, we instead focus on the nine real-world application datasets with real anomalies described above. A summarization of the key statistics of these datasets is presented in Table \ref{tab:stat}.

\begin{table}[t]
  \centering
  \caption{Key Statistics of Image Datasets. The first 15 datasets compose the MVTec AD dataset.}
  \scalebox{0.82}{
    \begin{tabular}{l@{}|ccc|cc}
    \hline
    \multirow{2}{3em}{  \textbf{Data} }   & \textbf{Original Training} & \multicolumn{2}{c|}{\textbf{Original Test}} & \multicolumn{2}{c}{\textbf{Anomaly Data}}\\
          \cline{2-6}
          & \textbf{Normal} & \textbf{Normal} & \textbf{Anomaly} & \textbf{\# Classes} & \textbf{Type} \\
          \hline
    Carpet & 280   & 28    & 89    & 5 & Texture \\
    Grid  & 264   & 21    & 57    & 5  & Texture\\
    Leather & 245   & 32    & 92    & 5 & Texture \\
    Tile  & 230   & 33    & 84    & 5  & Texture\\
    Wood  & 247   & 19    & 60    & 5  & Texture\\
    Bottle & 209   & 20    & 63    & 3  & Object\\
    Capsule & 219   & 23    & 109   & 5 & Object\\
    Pill  & 267   & 26    & 141   & 7 & Object\\
    Transistor & 213   & 60    & 40    & 4 & Object\\
    Zipper & 240   & 32    & 119   & 7 & Object\\
    Cable & 224   & 58    & 92    & 8& Object \\
    Hazelnut & 391   & 40    & 70    & 4 & Object\\
    Metal\_nut & 220   & 22    & 93    & 4 & Object\\
    Screw & 320   & 41    & 119   & 5 & Object\\
    Toothbrush & 60    & 12    & 30    & 1& Object \\
    \hline
    \textbf{MVTec AD} & 3,629 &	467	& 1,258 & 73 & -\\\hline
    \textbf{AITEX} & 1,692  & 564   & 183   & 12& Texture \\
    \textbf{SDD }  & 594   & 286   & 54    & 1 & Texture\\
    \textbf{ELPV}  & 1,131  & 377   & 715   & 2 & Texture\\
    \textbf{Optical} & 10,500 & 3,500  & 2,100  & 1& Object \\
    \textbf{Mastcam }& 9,302  & 426   & 451   & 11 & Object\\
    \textbf{BrainMRI} & 73    & 25    & 155   & 1 & Object\\
   \textbf{HeadCT} & 75    & 25    & 100   & 1 & Object\\
    \textbf{Hyper-Kvasir} & 2,021  & 674   & 757   & 4& Object \\
    \hline
    \end{tabular}%
    }
  \label{tab:stat}%
\end{table}%

\subsection{Competing Methods}\label{subsec:competing}

DevNet is compared with five state-of-the-art models from four closely related research lines, including two semi-supervised anomaly detection models: Deep SVDD (DSVDD) \cite{ruff2018deepsvdd} and KDAD \cite{salehi2021multiresolution}, weakly-supervised anomaly detection model: Deep SAD \cite{ruff2020deep}, imbalanced classification model: focal loss-driven classifier (FLOS) \cite{lin2017focalloss}, and multiple instance learning model: MINNS \cite{wang2018minns}.
\begin{itemize}
    \item \textbf{KDAD} is a deep anomaly detector based on multiresolution knowledge distillation. It trains the detection model by distilling the feature representations of a pre-trained model on ImageNet \cite{deng2009imagenet} at multiple layers, and detects anomalies based on the aggregation of the discrepancy between the features of the two models at these layers.
    \item \textbf{DSVDD} is a deep one-class classifier that imposes the SVDD objective \cite{tax2004svdd} on top of deep neural network-based representation to learn a one-class center-oriented hypersphere for describing a set of exclusively normal training samples. It considers samples that have large distances to the class center as anomalies.
    \item \textbf{Deep SAD} is an extension of DSVDD, which utilizes both labeled normal and abnormal samples to learn a more effective one-class description model. It adds a margin constraint into the original SVDD objective to enforce large distances between labeled anomalies and the one-class center and minimize the distances between normal data and the class center.
    \item \textbf{FLOS} is a deep imbalanced classifier that learns a binary classification model using the class-imbalance-sensitive loss -- focal loss \cite{lin2017focalloss}. It is further augmented by the popular imbalanced learning method -- batchwise oversampling \cite{he2009imbalance}, which is also used in DevNet, as described in Step 4 in Algorithm \ref{alg:devnet}.
    \item \textbf{MINNS} is a deep multiple instance classification model that applies multiple-instance-pooling layers \cite{wang2018revisiting} to intermediate and last feature layers to learn expressive bag features with deep neural networks. We also use the same oversampling method as in DevNet to enhance MINNS.
\end{itemize}

These contenders are selected mainly because (i) they are the recently proposed models achieving state-of-the-art performance on different benchmarks in diverse settings and (ii) their official implementation is made publicly available.

\subsection{Implementation Details}

The implementations of DSVDD\footnote{\url{https://github.com/lukasruff/Deep-SVDD-PyTorch}}, KDAD\footnote{\url{https://github.com/Niousha12/Knowledge_Distillation_AD}}, Deep SAD\footnote{\url{https://github.com/lukasruff/Deep-SAD-PyTorch}}, and MINNS\footnote{\url{https://github.com/yanyongluan/MINNs}} are taken from their authors. FLOS is implemented using exactly the same network architecture as DevNet, with the deviation loss function replaced with the focal loss. Particularly, DevNet first uses ResNet-18 without the fully connected layer as the backbone network to extract the feature map from the original image, and then learns the anomaly score of each entry in the feature map through a 1x1 convolutional layer, corresponding to the anomaly score of a patch of the original input image. It is followed by top-$K$ MIL-based anomaly score optimization using our deviation loss, in which each patch is treated as an instance. Lastly, the backbone and the anomaly score learning modules are trained in an end-to-end manner. By default, $K$ is set to 10\% of the size of the patch collection per image. DevNet also works well with other $K$ values (see Sec. \ref{subsec:ablation}).

For the optimization, we use Adam \cite{kingma2014adam} with an initial learning rate of $10^{-3}$, a weight decay of $10^{-2}$, and a batch size of 48. The network is trained for 50 epochs, which 20 iterations per epoch.

To exclude the effect of using different backbones in our comparison, Deep SAD and MINNS use the same backbone as DevNet, which enable them to perform better than using the original network. FLOS has exactly the same implementation as DevNet except the difference in the loss function. For DSVDD and KDAD, we keep the original network architecture, since they work less effectively with deeper networks like ResNet-18. Deep SAD, DSVDD and KDAD contains some specifically designed initialization and/or feature concatenation, so they are optimized with the recommended settings as in their original work. For FLOS and MINNS, we use the same optimization setting as DevNet.

\subsection{Performance Evaluation Metrics}

The popular detection performance metric, the Area Under Receiver Operating Characteristic Curve (AUC-ROC), is used in our experiments. AUC-ROC summarizes the ROC curve of true positives against false positives.
An AUC-ROC value of one indicates the best performance, while a value close to 0.5 indicates a random ranking of the examples. Larger AUC-ROC indicates better performance. AUC-ROC is widely used due to its good interpretability.
The reported AUC-ROC are averaged results over three independent runs with different random seeds, unless stated otherwise. For anomaly explanation, we mainly focus on the pixel-level AUC-ROC performance.

\subsection{Q1. Exploiting a Few Random Anomaly Examples}\label{subsec:random_anomalies}

\subsubsection{Experimental Setting}\label{subsubsec:randomexample}

The original training data contains exclusively normal data in the nine datasets used. Thus, to replicate the scenarios where we have a few random anomaly examples during training, we randomly sample a fixed number of anomaly samples from the test data to add into the training set as the labeled training anomaly examples. The remaining test data as our test set. Since only limited anomaly examples are available in real-world applications, the number of labeled anomalies is consistently fixed to 10 across all the nine datasets (see Sec. \ref{exp:labeledanomalies} for varying the number of anomaly examples). DevNet is compared with all the five state-of-the-art competing methods described in Sec. \ref{subsec:competing}.

\subsubsection{Results}

The AUC-ROC results of DevNet and its five competing methods are shown in Table \ref{tab:randomanomalies}. DevNet obtains the best performance on seven out of nine datasets, outperforming the best competing method per dataset by large margins on several datasets, \eg, AITEX (4.6\%), ELPV (2.8\%), and Hyper-Kvasir (5.6\%), with its performance close to the best performers on the other two datasets: Mastcam and HeadCT. For the 15 data subsets of MVTec AD, DevNet is the best performer on eight of them, achieving substantially better performance on some very challenging cases, \eg, Transistor and Screw; it achieves performance very close ($\leq0.2\%$) to the best contenders on three of the other datasets. As a result, DevNet achieves the best average AUC-ROC on MVTec AD, outperforming the competing methods by 0.6\%-21.8\%.

We also perform the paired \textit{Wilcoxon} signed rank test \cite{woolson2007wilcoxon} to examine the significance of the performance of DevNet against its competing methods across the (nine or 23) datasets. The p-value results are reported in Table \ref{tab:randomanomalies}, where `p-value (9)' are the results with MVTec AD considered as a whole while `p-value (23)' are the results with each data subset of MVTec AD treated separately. The `p-value (9)' results indicate significant improvement of DevNet over all its five contenders at the 95\% or 99\% confidence interval, while the `p-value (23)' results indicates the significant improvement at the 90\% or 95\% confidence interval.

Although the given anomaly examples are very limited, the `supervised' detectors -- DevNet, MINNS, FLOS and Deep SAD -- can utilize those labeled data to achieve substantially better performance than their unsupervised counterparts -- KDAD and DSVDD. This shows the importance of using labeled anomaly data in deep detection models, even when the amount of those data is limited. Compared to MINNS, FLOS and Deep SAD, DevNet learns more generalized representations of normality and abnormality by enforcing focal fine-grained normality representations and unbounded deviated abnormality representations, resulting in more superior performance on datasets with either object anomalies or texture anomalies.

\begin{table}[bt]
  \centering
  \caption{AUC-ROC Performance (mean±standard deviation) on Nine Image Datasets. The results for MVTec AD are the averaged results over its 15 data subsets.}
  \scalebox{0.75}{
    \begin{tabular}{l@{}p{1.2cm}p{1.2cm}p{1.2cm}p{1.2cm}p{1.2cm}c}
    \hline
    \textbf{Data} & \multicolumn{1}{c}{\textbf{KDAD}} & \multicolumn{1}{c}{\textbf{DSVDD}} & \multicolumn{1}{c}{\textbf{DeepSAD}} & \multicolumn{1}{c}{\textbf{FLOS}} & \multicolumn{1}{c}{\textbf{MINNS}} & \multicolumn{1}{c}{\textbf{DevNet}} \\\hline
    Carpet & 0.774±0.005 & 0.623±0.002 & 0.791±0.011 & 0.780±0.009 & \textbf{0.876}±0.015 & 0.867±0.040 \\
    Grid  & 0.749±0.017 & 0.915±0.001 & 0.854±0.028 & 0.966±0.005 & \textbf{0.983}±0.016 & 0.967±0.021 \\
    Leather & 0.948±0.005 & 0.800±0.008 & 0.833±0.014 & 0.993±0.004 & 0.993±0.007 & \textbf{0.999}±0.001 \\
    Tile  & 0.911±0.010 & 0.879±0.006 & 0.888±0.010 & 0.952±0.010 & 0.980±0.003 & \textbf{0.987}±0.005 \\
    Wood  & 0.940±0.004 & 0.895±0.009 & 0.781±0.001 & \textbf{1.000}±0.000 & 0.998±0.004 & 0.999±0.001 \\
    Bottle & 0.992±0.002 & 0.938±0.001 & 0.913±0.002 & \textbf{0.995}±0.002 & \textbf{0.995}±0.007 & 0.993±0.008 \\
    Capsule & 0.775±0.019 & 0.574±0.005 & 0.476±0.022 & 0.902±0.017 & \textbf{0.905}±0.013  & 0.865±0.057 \\
    Pill  & 0.824±0.006 & 0.691±0.001 & 0.875±0.063 & \textbf{0.929}±0.012 & 0.913±0.021 & 0.866±0.038 \\
    Transistor & 0.805±0.013 & 0.793±0.005 & 0.868±0.006 & 0.862±0.037 & 0.889±0.032 & \textbf{0.924}±0.027 \\
    Zipper & 0.927±0.018 & 0.687±0.004 & 0.974±0.005 & \textbf{0.990}±0.008 & 0.981±0.011 & \textbf{0.990}±0.009 \\
    Cable & 0.880±0.002 & 0.703±0.001 & 0.696±0.016 & 0.890±0.063 & 0.842±0.012 & \textbf{0.892}±0.020 \\
    Hazelnut & 0.984±0.001 & 0.769±0.010 & \textbf{1.000}±0.000 & \textbf{1.000}±0.000 & \textbf{1.000}±0.000 & \textbf{1.000}±0.000 \\
    Metal\_nut & 0.743±0.013 & 0.623±0.016 & 0.860±0.053 & 0.984±0.004 & 0.984±0.002 & \textbf{0.991}±0.006 \\
    Screw & 0.805±0.021 & 0.475±0.001 & 0.774±0.081 & 0.940±0.017 & 0.932±0.035 & \textbf{0.970}±0.015 \\
    Toothbrush & 0.863±0.029 & 0.941±0.016 & 0.885±0.063 & \textbf{0.900}±0.008 & 0.810±0.086 & 0.860±0.066 \\\hline
    \textbf{MVTec AD} & 0.861±0.009 & 0.727±0.001 & 0.830±0.009 & 0.939±0.007 & 0.939±0.011  & \textbf{0.945}±0.004 \\
    \textbf{AITEX} & 0.576±0.002 & 0.784±0.004 & 0.686±0.028 & 0.841±0.049 & 0.813±0.030 & \textbf{0.887}±0.013 \\
    \textbf{SDD}   & 0.888±0.005 & 0.682±0.001 & 0.963±0.005 & 0.967±0.018 & 0.961±0.016 & \textbf{0.988}±0.006 \\
    \textbf{ELPV} & 0.744±0.001 & 0.750±0.001 & 0.722±0.053 & 0.818±0.032 & 0.788±0.028 & \textbf{0.846}±0.022 \\
   \textbf{Optical} & 0.579±0.002 & 0.513±0.000 & 0.558±0.012 & 0.720±0.055 & 0.774±0.047 & \textbf{0.782}±0.065 \\
   \textbf{Mastcam} & 0.642±0.007 & 0.614±0.001 & 0.707±0.011 & 0.703±0.029 & \textbf{0.803}±0.031 & 0.790±0.021\\
    \textbf{BrainMRI} & 0.733±0.016 & 0.802±0.006 & 0.850±0.016 & 0.955±0.011 & 0.943±0.031 & \textbf{0.958}±0.012 \\
   \textbf{HeadCT} & 0.793±0.017 & 0.688±0.015 & 0.928±0.005 & 0.971±0.004 & \textbf{0.984}±0.010 & 0.982±0.009 \\
   \textbf{Hyper-Kvasir} & 0.401±0.002 & 0.515±0.000 & 0.719±0.032 & 0.773±0.029 & 0.647±0.051 & \textbf{0.829}±0.018 \\\hline
   \textbf{p-value (9)} & \centering 0.0039 & \centering 0.0039 & \centering 0.0039 & \centering 0.0039 & \centering 0.0391 & -\\
   \textbf{p-value (23)} & \centering 0.0000 & \centering 0.0000& \centering 0.0001 & \centering 0.0249 & \centering 0.0596 & - \\\hline
    \end{tabular}
    }
  \label{tab:randomanomalies}%
\end{table}%

\subsection{Q2. Open-set Anomaly Detection}

\subsubsection{Experimental Setting}

This section examines the detection performance under the open-set setting, where all anomaly classes in the test data are unseen during training. This generalization ability is important since anomalies are typically unbound to any specific class distribution. Particularly, we use the same procedure to create the training and test sets as that in Sec. \ref{subsubsec:randomexample}, except that the training anomaly examples are sampled from only one anomaly class and the test data contains the rest of other anomaly classes and the normal class. To create test data with unseen anomaly classes, this experiment is only applied to datasets having no less than two anomaly classes, including AITEX, ELPV, Mastcam and Hyper-Kvasir. Two challenging data subsets from MVTev AD, Carpet and Metal\_nut, are also used, which contain diverse texture/object anomaly classes of highly dissimilar appearance. For a given dataset with $k$ anomaly classes, exactly the same procedure is applied to each individual anomaly class, resulting in $k$ training-test sets with different seen and unseen anomaly classes. Some anomaly classes in AITEX and Mastcam contain only 1--5 samples, which are too small to train generalized models, so they are excluded in these experiments. As a result, we obtain 30 new datasets in total, with each dataset having exclusively unseen anomaly classes in its test set. DevNet is compared with three `supervised' models, Deep SAD, FLOS, and MINNS, with the best unsupervised model KDAD as baseline.

\subsubsection{Results}

The AUC-ROC results of DevNet and its four competing methods are shown in Table \ref{tab:openset}. DevNet is the best performer, achieving the best AUC-ROC on four tier-1 datasets and 12 tier-2 datasets, followed by KDAD that performs best on two tier-1 datasets and seven tier-2 datasets. The p-value results, which are based on the paired \textit{Wilcoxon} signed rank test across 30 tier-2 datasets, indicate the significant performance improvement of DevNet over all its four contenders at the 95\% confidence level.

Regardless of the open-set setting, supervised detectors like Deep SAD, FLOS and MINNS can still obtain quite good AUC-ROC results on some datasets, such as Carpet, Metal\_nut, AITEX and Hyper-Kvasir, indicating that they may learn some generalized representations of normality/abnormality. Compared to these detectors, our model gains substantially better generalizability on these datasets, boosting at least 5\% - 18\% AUC-ROC on many of the tier-2 datasets. On datasets where it is difficult to learn generalizable representations, such as Mastcam, the unsupervised detector KDAD performs better than anomaly-informed models, though all the models fail to achieve good performance on this dataset.

It is interesting to note that the representations learned using different anomaly classes are not mutually generalizable. That is, the detectors may gain substantially better performance having anomaly examples sampled from one class than from the other classes, see, for example, the AUC-ROC results on the dataset Mono \textit{vs.}\  that on Poly.

\begin{table}[bt]
  \centering
  \caption{AUC-ROC Performance under the Open-set Setting. Each dataset is named by the known anomaly class.}
  \scalebox{0.65}{
    \begin{tabular}{l@{}l@{}ccccc}
    \hline
    \textbf{Data (Tier-1)}      & \textbf{Data (Tier-2)} & \textbf{KDAD}  & \textbf{DeepSAD} & \textbf{FLOS}  & \textbf{MINNS} & \textbf{DevNet} \\
    \hline
    \multirow{6}[0]{*}{\textbf{Carpet}} & Color & \textbf{0.787}±0.005 & 0.736±0.007 & 0.760±0.005 & 0.767±0.011 & 0.767±0.015 \\
          & Cut   & 0.766±0.005 & 0.612±0.034 & 0.688±0.059 & 0.694±0.068 & \textbf{0.819}±0.037 \\
          & Hole  & 0.757±0.003 & 0.576±0.036 & 0.733±0.014 & 0.766±0.007 & \textbf{0.814}±0.038 \\
          & Metal & 0.836±0.003 & 0.732±0.042 & 0.678±0.083 & 0.789±0.097 & \textbf{0.863}±0.022 \\
          & Thread & 0.750±0.005 & 0.979±0.000 & 0.946±0.005 & \textbf{0.982}±0.008 & 0.972±0.009 \\\cline{2-7}
          & \textbf{Mean}  & 0.779 & 0.727 & 0.761 & 0.800 & \textbf{0.847} \\
    \hline
    \multirow{5}[0]{*}{\textbf{Metal\_nut}} & Bent  & 0.798±0.015 & 0.821±0.023 & 0.827±0.075 & 0.868±0.033 & \textbf{0.904}±0.022 \\
          & Color & 0.754±0.014 & 0.707±0.028 & 0.978±0.008 & \textbf{0.985}±0.018 & 0.978±0.016 \\
          & Flip  & 0.646±0.019 & 0.602±0.020 & 0.942±0.009 & \textbf{1.000}±0.000 & 0.987±0.004 \\
          & Scratch & 0.737±0.010 & 0.654±0.004 & 0.943±0.002 & 0.978±0.000 & \textbf{0.991}±0.017 \\\cline{2-7}
          & \textbf{Mean}  & 0.734 & 0.696 & 0.922 & 0.958 & \textbf{0.965} \\
    \hline
    \multirow{7}[0]{*}{\textbf{AITEX}} & Broken\_end & 0.552±0.006 & 0.442±0.029 & 0.585±0.037 & \textbf{0.708}±0.103 & 0.658±0.111 \\
          & Broken\_pick & \textbf{0.705}±0.003 & 0.614±0.039 & 0.548±0.054 & 0.565±0.018 & 0.585±0.028 \\
          & Cut\_selvage & 0.567±0.006 & 0.523±0.032 & \textbf{0.745}±0.035 & 0.734±0.012 & 0.709±0.039 \\
          & Fuzzyball & 0.559±0.008 & 0.518±0.023 & 0.550±0.082 & 0.534±0.058 & \textbf{0.734}±0.039 \\
          & Nep   & 0.566±0.006 & 0.733±0.017 & 0.746±0.060 & 0.707±0.059 & \textbf{0.810}±0.042 \\
          & Weft\_crack & 0.529±0.006 & 0.510±0.058 & \textbf{0.636}±0.051 & 0.544±0.183 & 0.599±0.137 \\\cline{2-7}
          & \textbf{Mean}  & 0.580 & 0.557 & 0.635 & 0.632 & \textbf{0.683} \\
    \hline
    \multirow{5}[0]{*}{\textbf{Hyper-Kvasir}} & Barretts & 0.405±0.003 & 0.666±0.012 & 0.764±0.066 & 0.608±0.064 & \textbf{0.834±0.012} \\
          & Barretts-short-seg & 0.404±0.003 & 0.672±0.017 & \textbf{0.810}±0.034 & 0.679±0.009 & 0.799±0.036 \\
          & Esophagitis-a & 0.435±0.002 & 0.619±0.027 & 0.815±0.022 & 0.665±0.045 & \textbf{0.844}±0.014 \\
          & Esophagitis-b-d & 0.367±0.003 & 0.564±0.006 & 0.754±0.073 & 0.480±0.043 & \textbf{0.810}±0.015 \\\cline{2-7}
          & \textbf{Mean}  & 0.403 & 0.630 & 0.786 & 0.608 & \textbf{0.822} \\
    \hline
    \multirow{3}[0]{*}{\textbf{ELPV}} & Mono  & \textbf{0.796}±0.002 & 0.554±0.063 & 0.629±0.072 & 0.557±0.010 & 0.599±0.040 \\
          & Poly  & 0.679±0.004 & 0.621±0.006 & 0.662±0.042 & 0.770±0.032 & \textbf{0.804}±0.022 \\\cline{2-7}
          & \textbf{Mean}  & \textbf{0.737} & 0.588 & 0.646 & 0.663 & 0.702 \\
    \hline
    \multirow{10}[0]{*}{\textbf{Mastcam}} & Bedrock & \textbf{0.638}±0.007 & 0.474±0.038 & 0.499±0.098 & 0.419±0.025 & 0.550±0.053 \\
          & Broken-rock & 0.590±0.007 & 0.497±0.054 & 0.608±0.085 & \textbf{0.687}±0.015 & 0.547±0.018 \\
          & Drill-hole & 0.630±0.006 & 0.494±0.013 & 0.601±0.009 & \textbf{0.651}±0.035 & 0.583±0.022 \\
          & Drt   & \textbf{0.711}±0.005 & 0.586±0.012 & 0.652±0.024 & 0.705±0.043 & 0.621±0.043 \\
          & Dump-pile & 0.697±0.007 & 0.565±0.046 & 0.700±0.070 & 0.697±0.022 & \textbf{0.705}±0.011 \\
          & Float & 0.632±0.007 & 0.408±0.022 & \textbf{0.736}±0.041 & 0.635±0.073 & 0.615±0.052 \\
          & Meteorite & \textbf{0.634}±0.007 & 0.489±0.010 & 0.568±0.053 & 0.551±0.018 & 0.554±0.021 \\
          & Scuff & \textbf{0.638}±0.006 & 0.502±0.010 & 0.575±0.042 & 0.502±0.040 & 0.528±0.034 \\
          & Veins & \textbf{0.621}±0.007 & 0.542±0.017 & 0.608±0.044 & 0.577±0.013 & 0.589±0.072 \\\cline{2-7}
          & \textbf{Mean}  & \textbf{0.644} & 0.506 & 0.616 & 0.603 & 0.588 \\
    \hline
    & \textbf{p-value} & 0.0144 & 0.0000 & 0.0341 & 0.0201 & - \\\hline
    \end{tabular}%
    }
  \label{tab:openset}%
\end{table}%

\subsection{Q3. Anomaly Explanation}\label{subsec:explanation}

\subsubsection{Experimental Setting}
This section performs quantitative and qualitative evaluation of the DevNet's explainability. This is done by looking at how effectively it localizes the features that explain why the images are detected as anomalies. The experiments are based on the widely-used dataset -- MVTec AD -- that provides the pixel-level ground truth of anomaly masks. We also present a qualitative analysis with the visualization of our anomaly localization of diverse types of anomaly.

 \subsubsection{Results}

The pixel-level AUC-ROC performance of DevNet in anomaly localization is shown in Table \ref{tab:explanation}, with KDAD, FLOS, and MINNS as baseline. It is clear that DevNet outperforms all three competing methods, providing more accurate localization of the cause of the anomalies for both of the texture and object anomalies on highly diverse visual surfaces. The results in Table \ref{tab:explanation} are based on all the anomaly samples in the test data, without involving the anomaly detection step. However, in practice, it is important to be accurate in both of the detection and localization steps. 

\begin{table}[b]
  \centering
  \caption{Pixel-level AUC-ROCs in Anomaly Localization}
  \scalebox{0.85}{
    \begin{tabular}{llcccc}
    \hline
    & \textbf{Dataset}     & \textbf{KDAD} & \textbf{FLOS} & \textbf{MINNS} & \textbf{DevNet} \\\hline
   \multirow{5}[0]{*}{\textbf{Texture}} & Carpet & 0.956 & \textbf{0.979} & 0.887 & 0.963 \\
    & Grid  & 0.918 & \textbf{0.978} & 0.647 & 0.935 \\
   &  Leather & 0.981 & 0.988 & 0.865 & \textbf{0.990} \\
   & Tile  & 0.828 & 0.914 & 0.858 & \textbf{0.950} \\
   & Wood  & 0.848 & \textbf{0.913} & 0.663 & 0.900 \\\hline
   \multirow{10}[0]{*}{\textbf{Object}} & Bottle & \textbf{0.963} & 0.892 & 0.781 & 0.951 \\
    & Capsule & \textbf{0.959} & 0.951 & 0.951 & 0.938 \\
    & Pill  & 0.896 & \textbf{0.909} & 0.800 & 0.859 \\
    & Transistor & 0.765 & 0.862 & 0.800 & \textbf{0.839} \\
    & Zipper & 0.939 & \textbf{0.974} & 0.829 & 0.973 \\
    & Cable & 0.824 & 0.913 & 0.851 & \textbf{0.920} \\
    & Hazelnut & 0.946 & 0.935 & 0.699 & \textbf{0.959} \\
    & Metal\_nut & 0.864 & 0.835 & 0.835 & \textbf{0.876} \\
    & Screw & \textbf{0.960} & 0.859 & 0.793 & 0.897 \\
    & Toothbrush & \textbf{0.961} & 0.743 & 0.752 & 0.819 \\ \hline
    & \textbf{Mean}  & 0.907 & 0.910 & 0.801 & \textbf{0.918} \\\hline
    \end{tabular}%
    }
  \label{tab:explanation}%
\end{table}%

\begin{figure}[h!]
  \centering
    \includegraphics[width=0.45\textwidth]{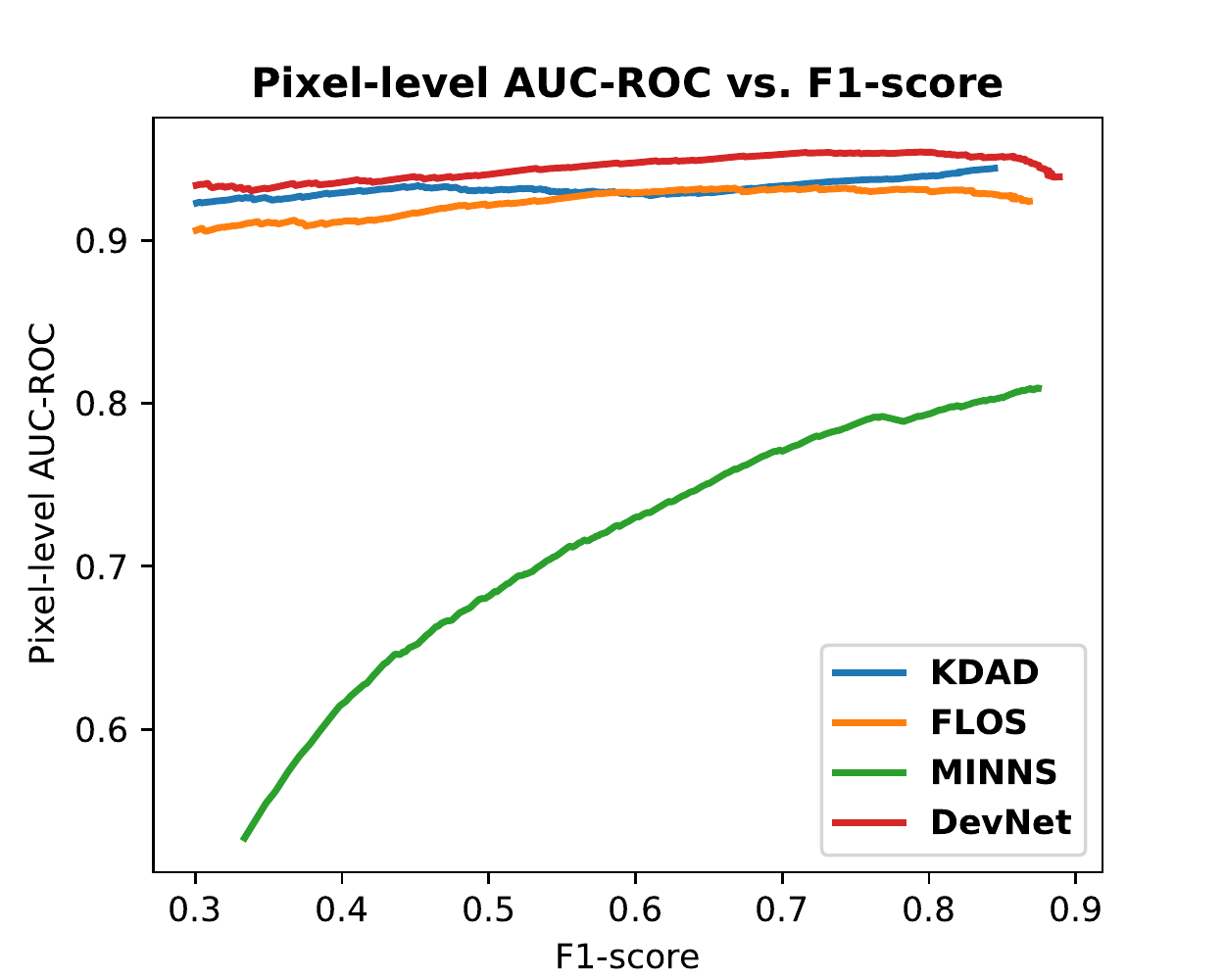}
  \caption{AUC-ROC vs.\  F1-score Curve in Anomaly Localization.}
  \label{fig:auc_fscore}
\end{figure}

To unify the evaluation of detection and localization steps, we present the results of both steps using a curve of pixel-level AUC-ROC (localization) vs. F1-score (detection) in Fig. \ref{fig:auc_fscore}. Different from the results in Table \ref{tab:explanation}, the pixel-level AUC-ROC here is based on the true anomalies detected by the model rather than all the anomalies in the test data, while F1-score is calculated based on a specific cut-off decision threshold. We use a large number of incrementally increased cut-off thresholds for each model to obtain smooth curve results. Both AUC-ROC and F1-score in Fig. \ref{fig:auc_fscore} are averaged results over all the 15 anomaly classes of MVTec AD. These results show that DevNet consistently outperforms all of its three counterparts in the joint anomaly detection and localization evaluation. Particularly, DevNet is able to achieve F1-score close to 90\% while maintaining an AUC-ROC of about 95\%, which is a very promising result for real-world applications. KDAD and MINNS make aggressive anomaly localization, which can often well localize the anomalous regions but wrongly identify large normal regions as anomalous features, leading to much higher false negative errors than DevNet.

\begin{figure*}[t!]
  \centering
    \includegraphics[width=0.95\textwidth]{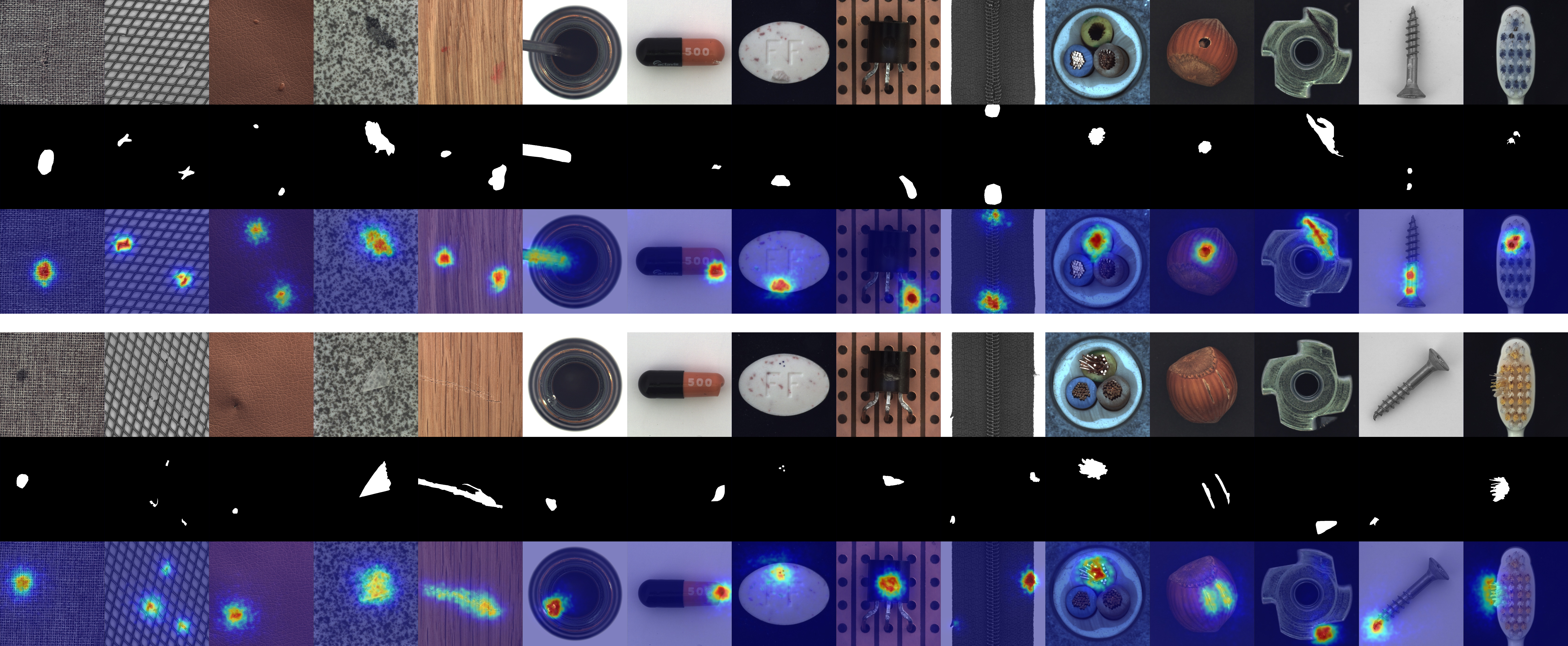}
  \caption{Visualization of Anomaly Localization for Two Groups of Images in MVTec AD. For each group of results, the first row is input image, followed by the pixel-level ground truth -- anomaly mask -- and our anomaly localization result. For each of the 15 anomaly classes, we present visualization of exemplar images from two different fine-grained types of anomaly within the class: Carpet (Thread and Color), Grid (Bent and Broken), Leather (Glue and Poke), Tile (Gray Stroke and Glue Strip), Wood (Color and Scratch), Bottle (Contamination and Broken Small), Capsule (Crack and Squeeze), Pill (Contamination and Color), Transistor (Cut Lead and Damaged Case), Zipper (Split Teeth and Fabric Border), Cable (Missing Wire and Bent Wire), Hazelnut (Hole and Cut), Metal\_nut (Color and Bent), Screw (Scratch Neck and Manipulated Front), and Toothbrush (no fine-grained anomaly types for this case).}
  \label{fig:localization}
\end{figure*}

The visualization of our anomaly localization results is presented in Fig. \ref{fig:localization}. The results show that DevNet can accurately localize the anomalous pixels of different anomaly classes on different surfaces, regardless of the size or shape of the anomalous regions, offering straightforward explanation of the identified anomalies. 

\subsection{Further Empirical Analysis of DevNet }\label{subsec:ablation}

This section presents a thorough analysis of DevNet from different aspects to gain deep insights into its capabilities.

\subsubsection{Q4. Ablation Study}
DevNet includes three key modules, including network backbone for feature representations, deviation-based anomaly score learning, and MIL-based fine-grained feature learning. We compare DevNet to its five variants to evaluate the contribution of each of these modules.

\begin{itemize}
    \item \textbf{Rep}. We devise a variant of DevNet, called Rep, which removes the anomaly score learning layer (\ie, the output layer) and uses our deviation loss to learn the representations only. In this case, the reference in the loss function is drawn from a multivariate Gaussian rather than a univariate Gaussian.
    \item \textbf{DS}. DS is DevNet without having fine-grained features, \ie, DevNet that has a scalar output layer but does not use top-$K$ multiple instance learning.
    \item \textbf{Patch-DS}. We also derive Patch-DS that applies deviation loss to every patch of a given image and calculates the anomaly scores by averaging the scores over all the image patches rather than top-$K$ MIL.
    \item \textbf{AlexNet + DS} and \textbf{ResNet-50 + DS}. ResNet-18 is the default backbone, which is used in our DevNet model (denoted as MIL-DS) and the three variants described above. We also design a shallower model, AlexNet + DS, and a deeper model, ResNet-50 + DS.
\end{itemize}

The ablation study results are shown in Table \ref{tab:ablation_image}. It is clear that DS significantly improves the performance of Rep across the datasets, indicating the remarkable contribution of enforcing differentiable anomaly score learning. Directly enforcing deviation loss on each individual image patch independently in Patch-DS helps gain some improvement on some datasets, such as Bottle and Transistor, compared to DS, but it severely degrades the performance on most datasets. This is mainly because there are some fine-grained features DS fail to capture, but learning deviation network across all image patches is inevitably biased by irrelevant image patches. By contrast, our proposed model MIL-DS learns to select and optimize on the most relevant image patches by top-$K$ multiple-instance deviation learning, outperforming both DS and Patch-DS by large margins on most datasets while still maintaining good performance on a few exceptional datasets like Pill, Transistor and Toothbrush. This highlights the great importance of the MIL component. Additionally, as shown by the results of AlexNet + DS, the use of a shallower backbone may help DevNet gains better performance on some datasets such as Capsule, but degrades its performance on most datasets. ResNet-50 + DS outperforms ResNet-18 + DS on only a few datasets, whereas the reverse cases occur on most datasets, indicating that using a much deeper backbone -- ResNet-50 -- is mostly not beneficial, partly due to the lack of large anomaly data. 

\begin{table}[htbp]
  \centering
  \caption{AUC Results of the Ablation Study on Image Datasets. MIL-DS with ResNet-18 is the default DevNet model. The others are its variants.}
  \scalebox{0.75}{
    \begin{tabular}{p{1.35cm}p{1.2cm}p{1.2cm}p{1.25cm}p{1.3cm}|p{1.2cm}c}
    \hline
    \textbf{Module}   & \centering \textbf{Rep} & \centering \textbf{DS} & \centering \textbf{Patch-DS} & \centering \textbf{MIL-DS} & \multicolumn{2}{c}{\textbf{DS}} \\ \hline
    $L_{\mathit{dev}}$ & \centering$\checkmark$ &\centering $\checkmark$ &\centering $\checkmark$ & \centering $\checkmark$& \multicolumn{2}{c}{$\checkmark$} \\
    $\eta$ &  & \centering$\checkmark$ & \centering$\checkmark$& \centering$\checkmark$& \multicolumn{2}{c}{$\checkmark$} \\ 
    \textbf{Patchwise} &  & &\centering $\checkmark$ & & \multicolumn{2}{c}{} \\
    $L_{\mathit{mil}}$ &  & & & \centering $\checkmark$& \multicolumn{2}{c}{} \\ \hline
     \textbf{Data}   & \multicolumn{4}{c|}{\textbf{ResNet-18}} & \multicolumn{1}{c}{\textbf{AlexNet}} & \multicolumn{1}{c}{\textbf{ResNet-50}} \\ \hline
    Carpet & 0.760±0.030 & 0.696±0.124 &  0.614±0.062 & \textbf{0.867}±0.040 & 0.347±0.031 & 0.678±0.122 \\
    Grid & 0.963±0.014 & 0.956±0.038  & \textbf{0.981}±0.005 & 0.967±0.021 & 0.956±0.042 & 0.979±0.009 \\
    Leather& 0.994±0.011 & 0.992±0.008  & 0.948±0.056 & 0.999±0.001 & \textbf{1.000}±0.000 & 0.776±0.119 \\
    Tile  & 0.788±0.008 & 0.980±0.010  & \textbf{0.987}±0.007 & \textbf{0.987}±0.005 & 0.897±0.033 & 0.950±0.062 \\
    Wood & 0.994±0.004 & 0.996±0.007  & \textbf{1.000}±0.000 & 0.999±0.001 & 0.967±0.036 & 0.994±0.007 \\
    Bottle & 0.943±0.019 & 0.976±0.021 & \textbf{0.997}±0.007 & 0.993±0.008 & 0.979±0.006 & 0.985±0.014 \\
    Capsule & 0.636±0.089 & 0.720±0.104 & 0.759±0.063 & 0.865±0.057 & \textbf{0.881}±0.017 & 0.776±0.077 \\
    Pill & 0.804±0.098 & \textbf{0.922}±0.014  & 0.921±0.018 & 0.866±0.038 & 0.896±0.033 & 0.887±0.072 \\
    Transistor & 0.759±0.038 & 0.888±0.044  & \textbf{0.956}±0.013 & 0.924±0.027 & 0.807±0.015 & 0.891±0.060 \\
    Zipper  & 0.988±0.005 & 0.972±0.011 & 0.982±0.010 & \textbf{0.990}±0.009 & 0.937±0.016 & 0.952±0.032 \\
    Cable & 0.724±0.023 & 0.874±0.028  & 0.869±0.021 & \textbf{0.892}±0.020 & 0.868±0.030 & 0.836±0.060 \\
    Hazelnut & \textbf{1.000}±0.000 & \textbf{1.000}±0.000 & \textbf{1.000}±0.000 & \textbf{1.000}±0.000 & 0.987±0.007 & \textbf{1.000}±0.000 \\
    Metal\_nut & 0.942±0.030 & 0.962±0.026  & 0.987±0.004 & \textbf{0.991}±0.006 & 0.961±0.007 & 0.958±0.035 \\
    Screw & 0.887±0.053 & \textbf{0.971}±0.012  & 0.953±0.025 & 0.970±0.015 & 0.903±0.017 & 0.961±0.032 \\
    Toothbrush & 0.658±0.056 & \textbf{0.913}±0.048  & 0.895±0.082 & 0.860±0.066 & 0.860±0.066 & 0.837±0.085 \\\hline
    \textbf{MVTecAD}& 0.856±0.010 & 0.921±0.013  & 0.923±0.015 & \textbf{0.945}±0.004 & 0.883±0.009 & 0.897±0.017 \\
    \textbf{AITEX} & 0.860±0.017 & 0.853±0.035  & 0.796±0.053 & \textbf{0.887}±0.013 & 0.743±0.025 & 0.859±0.016 \\
    \textbf{SDD}  & 0.865±0.054 & 0.982±0.009  & 0.949±0.036 & \textbf{0.988}±0.006 & 0.974±0.010 & 0.984±0.009 \\
    \textbf{ELPV} & 0.755±0.012 & 0.820±0.024  & 0.728±0.053 & \textbf{0.846}±0.022 & 0.705±0.048 & 0.828±0.019 \\
    \textbf{optical}  & 0.777±0.007 & 0.769±0.044 & 0.776±0.063 & \textbf{0.782}±0.065 & 0.622±0.026 & 0.766±0.029 \\
    \textbf{BrainMRI} & 0.851±0.029 & 0.964±0.008 & \textbf{0.969}±0.009 & 0.958±0.012 & 0.912±0.027 & 0.904±0.059 \\
    \textbf{HeadCT} & 0.876±0.035 & 0.967±0.015  & 0.963±0.021 & \textbf{0.982}±0.009 & 0.934±0.019 & \textbf{0.982}±0.009 \\
    \textbf{Hyperkvasir} & 0.602±0.015 & 0.759±0.056  & 0.575±0.048 & \textbf{0.829}±0.018 & 0.696±0.013 & 0.812±0.038 \\\hline
    \end{tabular}%
    }
  \label{tab:ablation_image}%
\end{table}%

\begin{figure}[h!]
  \centering
    \includegraphics[width=0.485\textwidth]{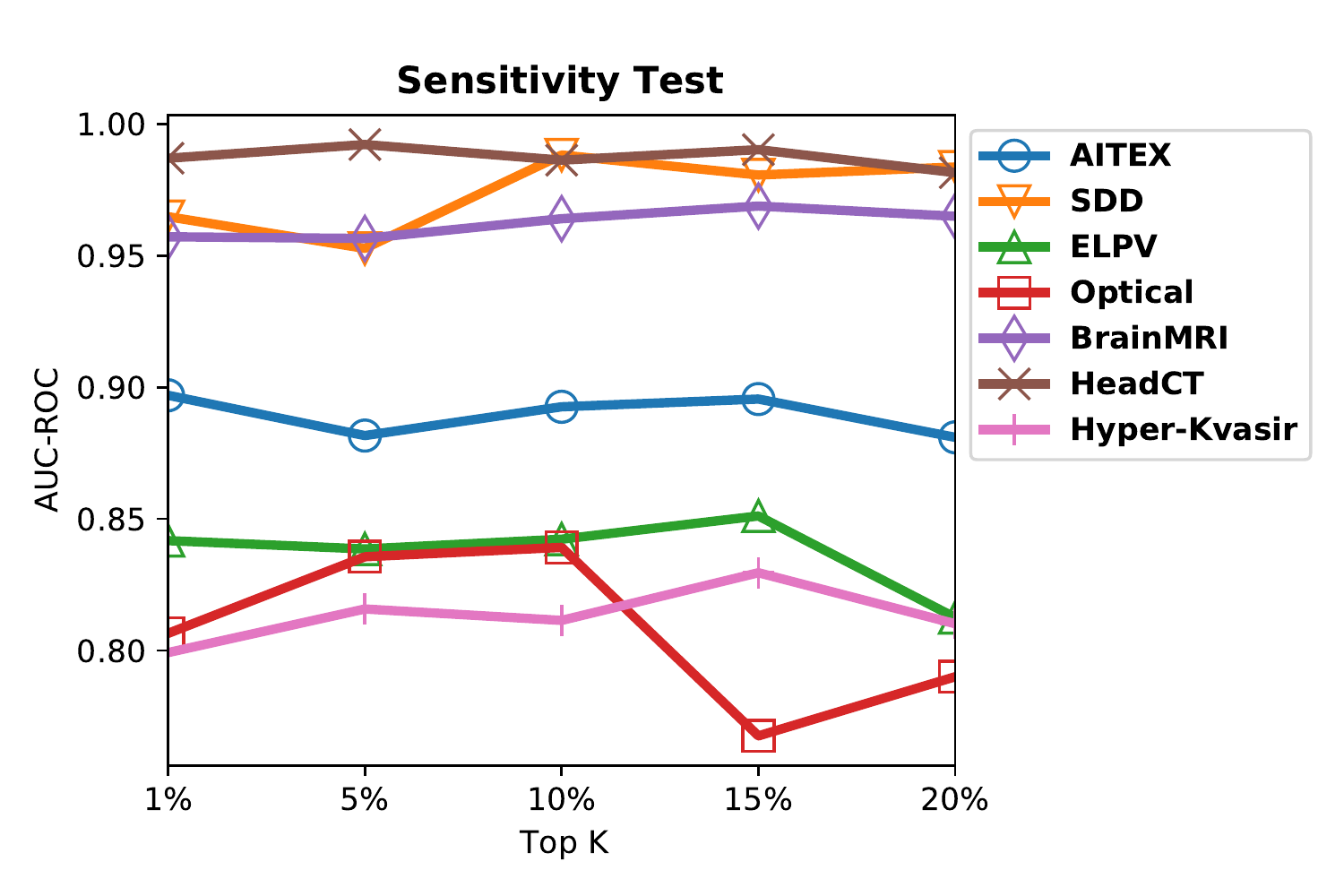}
  \caption{AUC-ROC \textit{w.r.t.} Different $K$ Values. Mastcam is excluded because the input resolution ($64 \times 64$) is too small to use large $K$.}
  \label{fig:robustness_K}
\end{figure}

We further look into the key module -- top-$K$ multiple-instance deviation learning -- by examining the use of different $K$ values. The results are reported in Fig. \ref{fig:robustness_K}. DevNet generally obtains better performance with $K$ increasing from 1\% to 10\% across all the datasets; it then starts to decline when $K=15\%$ or $K=20\%$ on some datasets. $K=10\%$ is generally recommended, as it enables DevNet to obtain the (nearly) best performance on almost all the datasets used.

\subsubsection{Q5. Sample Efficiency}\label{exp:labeledanomalies}
This section examines the sample efficiency of DevNet, with three anomaly-informed models Deep SAD, FLOS, and MINNS as baseline. Utilizing labeled sample efficiently is a critical capability,  as it is difficult to obtain large labeled anomaly data in real-world applications. The number of available labeled anomalies in training data varies from 5 to 40, with the test data unchanged. This experiment is based on the same experimental protocol as in Sec. \ref{subsubsec:randomexample}, and it is performed on all our datasets except MVTec AD. 

\begin{figure*}[h!]
  \centering
    \includegraphics[width=0.95\textwidth]{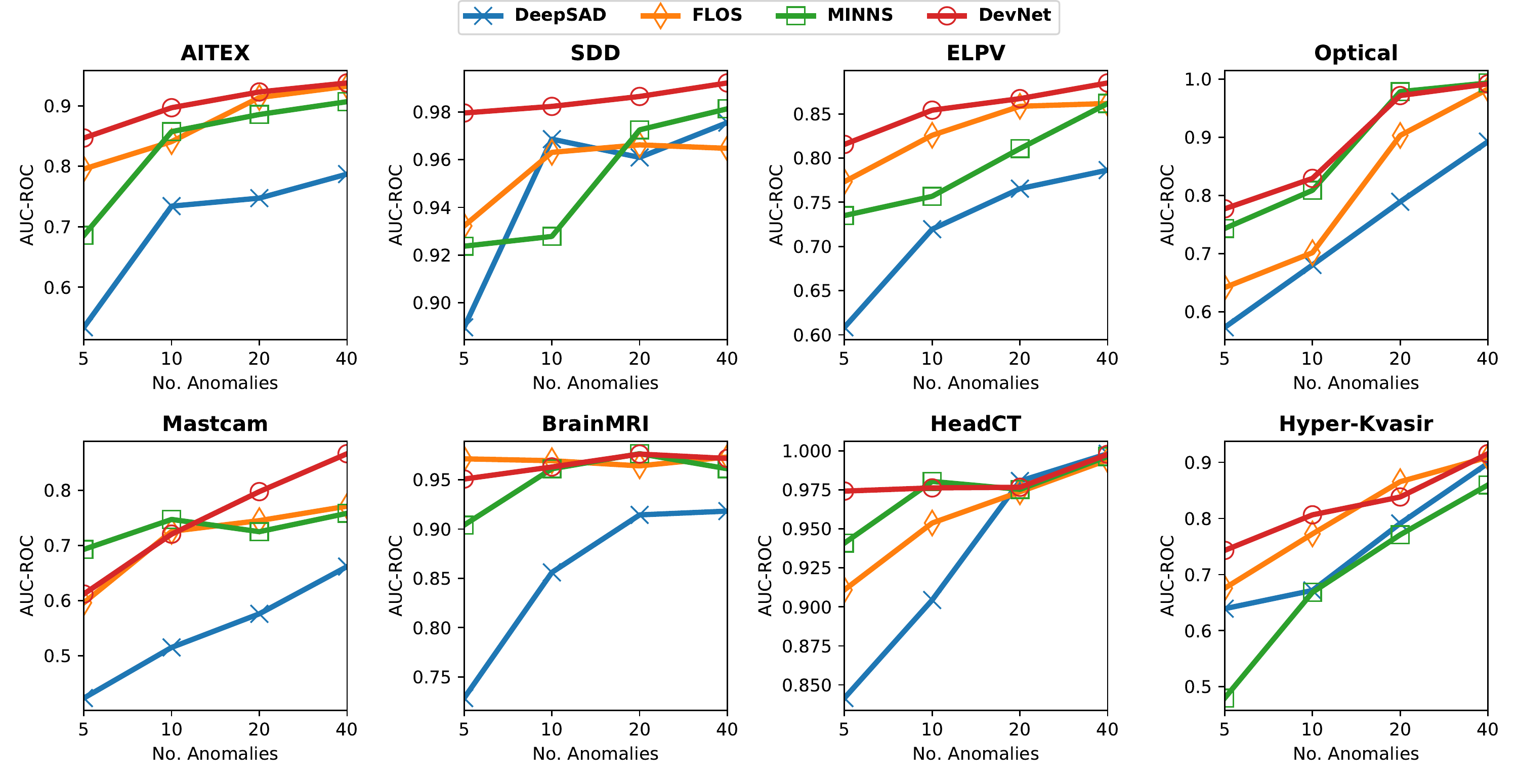}
  \caption{AUC-ROC Performance \textit{w.r.t.} Different Number of Labeled Anomaly Examples Used.}
  \label{fig:sample_efficiency}
\end{figure*}

The AUC-ROC results are reported in Fig. \ref{fig:sample_efficiency}. All four methods obtains better performance within increasing amount of labeled anomaly data. Consistent to the results in Table \ref{tab:randomanomalies}, DevNet is consistently the best performer on most datasets with different number of labeled anomalies, and it performs comparably good to the best performer otherwise. Impressively, even when using 50\% - 75\% less labeled data, DevNet is able to outperform the strong competing method MINNS by a large margin on many of these datasets, \eg, DevNet using 10 or 20 labeled anomalies vs. MINNS using 40 labeled anomalies on AITEX, SDD, ELPV, Mastcam and Hyper-Kvasir. DevNet gains similar, or larger, sample efficiency improvement compared to the other two counterparts, \ie, FLOS on SDD, ELPV, Optical, and Mastcam; Deep SAD on AITEX, SDD, ELPV, Optical, Mastcam, and BrainMRI. It is clear that the fine-grained feature learning is one key driving force for the performance of DevNet as the similar method -- MINNS -- also achieves better results than the other counterparts in many cases. Further, the end-to-end differentiable score learning enables DevNet to exploit the limited anomaly examples in a more direct way, resulting in more sample-efficient learning.

\subsubsection{Q6. Robustness w.r.t. Anomaly Contamination}

This section investigates the robustness of DevNet \textit{w.r.t.} different anomaly contamination levels in the training normal data, with Deep SAD, FLOS and MINNS as baseline. We vary the contamination level from 0\% up to 20\%, with the number of available labeled anomalies fixed to be 10. Particularly, the training normal data is polluted by anomaly images randomly sampled from the test data; and this sampled anomaly data is then removed from the test data. The experiment is focused on ELPV, BrainMRI, HeadCT and Hyper-Kvasir, as the other datasets do not contain sufficiently large anomaly data in the test set for the use of anomaly pollution.

\begin{figure}[h!]
  \centering
    \includegraphics[width=0.485\textwidth]{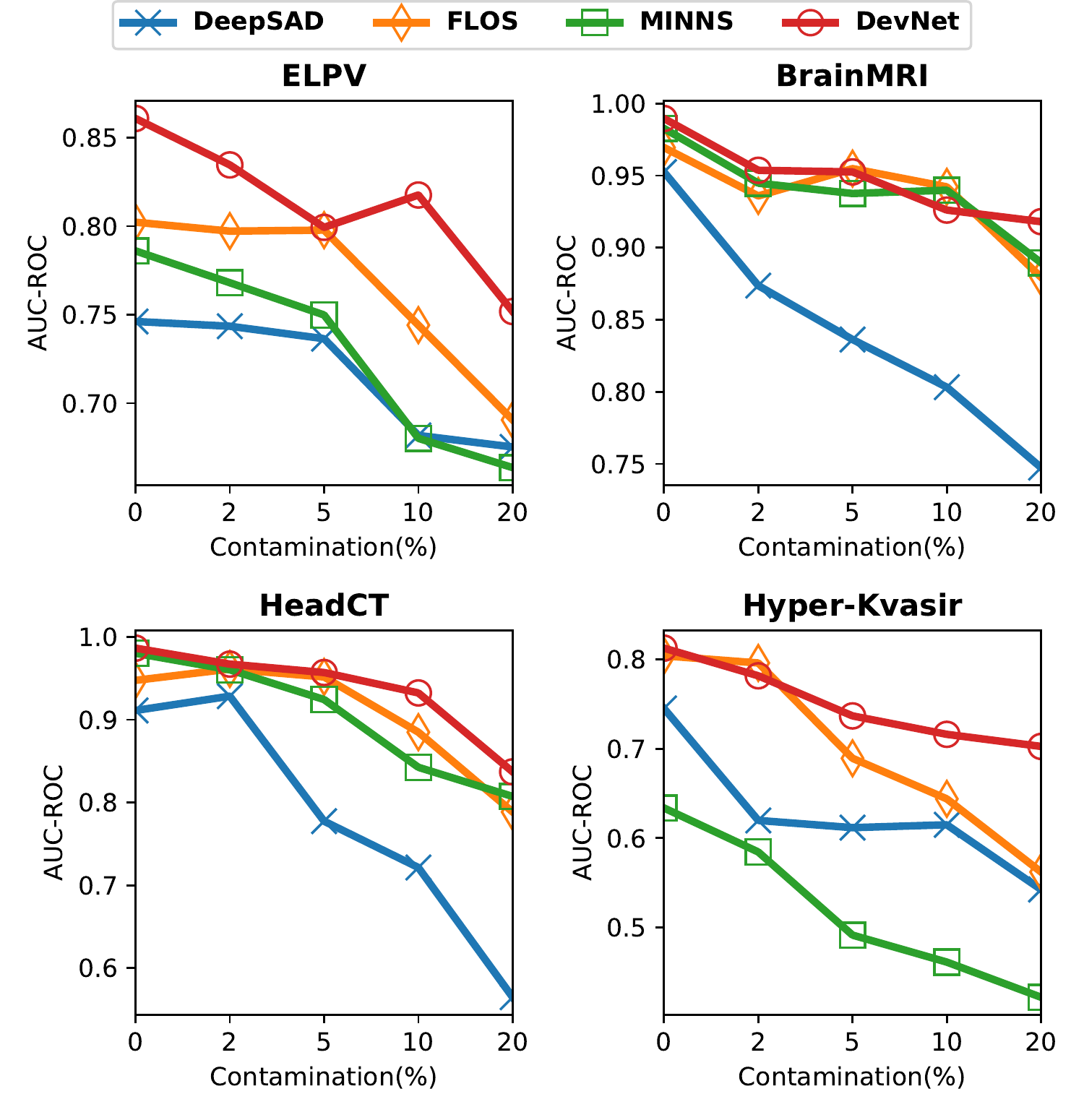}
  \caption{AUC-ROC \textit{w.r.t.} Anomaly Contamination Rate.}
  \label{fig:image_robustness}
\end{figure}

The AUC-ROC results are shown in Fig. \ref{fig:image_robustness}. As expected, the performance of all four detectors declines with increasing contamination rate. DevNet is more robust to the contamination, decreasing at a slower rate than the three contenders. Importantly, the improvement of DevNet over its contenders becomes clearer at large contamination rates such as at rate of 10\% or 20\%. This superiority benefits from the score-level Gaussian prior-driven learning that allows more flexible representations of normality.

\section{Conclusions and Future Work}

This paper considers a few-shot anomaly detection (FSAD) problem and introduces a novel framework and its instantiation DevNet to tackle the problem. This is achieved by leveraging a few labeled anomalies with a Gaussian prior to fulfill an end-to-end differentiable learning of fine-grained anomaly scores. By doing so, DevNet can be trained more robustly \textit{w.r.t.} anomaly contamination and more sample-efficiently, and learns substantially improved generalized feature representations. This allows significantly better AUC-ROC performance in both anomaly detection and explanation compared to state-of-the-art semi/weakly-supervised anomaly detectors and fully supervised classifiers on nine real-world image anomaly detection datasets. 

We also treats FSAD as an open-set problem and evaluate the detectors under the open-set settings. Our theoretical and empirical analysis results show that DevNet is able generalized to the open-set setting and achieves significantly better performance than the aforementioned counterparts.

It is generally difficult to achieve the best performance of an anomaly detector on each specific dataset, as shown by the results of DevNet and its variants in Table \ref{tab:ablation_image}. We plan to investigate automated machine learning techniques to automatically choose the network architectures and the relevant modules to further enhance DevNet and other state-of-the-art detectors. It would be also important to explore more advanced techniques to further reduce both the empirical risk and open space risk as in Eqn. (\ref{eqn:bothrisk}) for more effective `supervised' open-set anomaly detection.

\section*{Acknowledgments}

Part of the method presented here appeared in
\cite{pang2019devnet}, where we only focus on anomaly detection in tabular data.

\bibliographystyle{IEEEtran}
\bibliography{references}

\end{document}